\documentclass[11pt]{article}

\usepackage[final]{acl}

\usepackage{times}
\usepackage{latexsym}

\usepackage{caption}
\usepackage{placeins}

\usepackage{cuted}

\usepackage{tikz}
\usetikzlibrary{positioning,arrows.meta}
\usepackage[T1]{fontenc}

\usepackage[utf8]{inputenc}

\usepackage{microtype}

\usepackage{inconsolata}

\usepackage{graphicx}

\usepackage[many]{tcolorbox}
\usepackage{subcaption}
\usepackage{graphicx}
\usepackage{adjustbox} 
\usepackage{multirow}
\usepackage{booktabs}
\usepackage[table]{xcolor}
\definecolor{lightteal}{RGB}{230,245,245}
\definecolor{lightblue}{RGB}{220,235,250}
\definecolor{lightgreen}{RGB}{235,250,235}
\definecolor{lightred}{RGB}{254, 216, 177}
\definecolor{fem}{HTML}{F6D4FF}
\definecolor{masc}{HTML}{ADF7C7}
\definecolor{neut}{HTML}{f8ffe3}

\definecolor{incStrong}{RGB}{169,208,142}
\definecolor{incexStrong}{RGB}{112,173,71}
\definecolor{incWeak}{RGB}{226,239,218}
\definecolor{same}{RGB}{230,230,230}
\definecolor{decWeak}{RGB}{252,228,214}
\definecolor{decStrong}{RGB}{244,177,131}
\title{From Gender Priors to Contextual Disambiguation in MT:\\Causal Evidence of Gender Encoding in Attention Heads}

\author{
\textbf{Chiara Manna}\textsuperscript{1} \quad
\textbf{Hosein Mohebbi}\textsuperscript{2} \quad
\textbf{Eva Vanmassenhove}\textsuperscript{1} \\
\textbf{Frédéric Blain}\textsuperscript{1} \quad
\textbf{Afra Alishahi}\textsuperscript{1} \\
\textsuperscript{1}Tilburg University
\qquad
\textsuperscript{2}University of Amsterdam \\
\texttt{\{c.manna,e.o.j.vanmassenhove,f.l.g.blain,a.alishahi\}@tilburguniversity.edu} \\
\texttt{h.mohebbi@uva.nl}
}

\begin{document}

\newtcolorbox{PromptBox}[1][]{width=\columnwidth,
  colback=gray!8, colframe=gray!8,
  colbacktitle=gray!30,
  boxrule=0.6pt, arc=6pt,
  left=2pt, right=2pt, top=6pt, bottom=6pt,
  fontupper=\ttfamily\small,
  title=#1
}

\newtcbox{\malehl}{on line, boxrule=0pt, arc=2pt,
  colback=masc, colframe=masc, left=2pt, right=2pt, top=0.2ex, bottom=0.2ex}
\newtcbox{\femalehl}{on line, boxrule=0pt, arc=2pt,
  colback=fem,  colframe=fem,  left=2pt, right=2pt, top=0.2ex, bottom=0.2ex}

\newtcolorbox{LangBox_correct}{
  width=\columnwidth,
  colback=lightteal,
  colframe=lightteal,
  arc=4pt,
  boxrule=0.4pt,
  left=4pt, right=4pt, top=3pt, bottom=3pt,
  fontupper=\footnotesize\centering,
  before skip=2pt, after skip=2pt,
  boxsep=2pt
}

\newtcolorbox{LangBox_incorrect}{
  width=\columnwidth,
  colback=lightred,
  colframe=lightred,
  arc=4pt,
  boxrule=0.4pt,
  left=4pt, right=4pt, top=3pt, bottom=3pt,
  fontupper=\footnotesize\centering,
  before skip=2pt, after skip=2pt,
  boxsep=2pt
}
\maketitle
\begin{abstract}
Machine translation offers a natural testbed for studying gender bias by making gender-related modeling decisions directly observable in translation outputs. When translating from languages with sparse gender marking into morphologically richer ones, correct gender realizations often depend on contextual cues (e.g., \textit{he}/\textit{she}) and require models to integrate context into their decision process. While recent work has shown that MT-specific post-training can improve contextual sensitivity at the output level, the internal mechanisms underlying these improvements remain unclear. We address this gap through a targeted attention-head ablation analysis in state-of-the-art MT-tuned Large Language Models.  
Our results provide causal evidence that gender disambiguation is mediated by a small subset of attention heads that are particularly responsive to feminine signals. Ablating them disproportionately disrupts feminine realizations, while masculine ones remain largely unaffected. This asymmetry is consistent with a correction mechanism, in which explicit contextual grounding is required to override a masculine default.\footnote{Code and data available at \url{https://github.com/chiaramanna/gender-heads-mt}}
\end{abstract}

\section{Introduction}
Gender bias in Machine Translation (MT) has been extensively documented over the past decade \citep{savoldi2025decade}, alongside broader concerns about systematic biases in Natural Language Processing (NLP) \citep{sun-etal-2019-mitigating,costa2019analysis,blodgett-etal-2020-language,stanczak2021survey}. Due to its cross-linguistic nature, MT constitutes a particularly challenging and therefore informative setting in which biases can both emerge and be exposed. As a matter of fact, translating from languages where gender marking is absent or sparse (e.g., Hungarian or English) into morphologically richer ones (e.g., Italian, German or Hindi) requires implicit gender information to be made explicit in the target \citep{stahlberg,ackerman,cao-daume-iii-2020-toward,vanmassenhove-etal-2018-getting,vanmassenhove2024gender}. This process, known as gender disambiguation, can reveal a model's tendency to rely on statistical patterns observed in the training data, often resulting in stereotypical errors or default-to-masculine realizations that reflect a male-skewed distribution in the training corpora \citep{vanmassenhove-hardmeier-2018-europarl}. 

\begin{figure}[h!]
\centering

\begin{LangBox_correct}
The cook prepared a soup for the \textbf{housekeeper} because
\malehl{\textbf{he}} helped clean the room.
\end{LangBox_correct}

\begin{LangBox_correct}
Il cuoco ha preparato una zuppa per
\femalehl{\textbf{la governante}} perché aveva aiutato a pulire la stanza.
\end{LangBox_correct}

\vspace{0.25em}

\begin{LangBox_correct}
The cook prepared a soup for the \textbf{housekeeper} because
\femalehl{\textbf{she}} helped clean the room.
\end{LangBox_correct}

\begin{LangBox_correct}
Il cuoco ha preparato una zuppa per
\femalehl{\textbf{la governante}} perché lei aveva aiutato a pulire la stanza.
\end{LangBox_correct}

\vspace{0.5em}
\caption[]{
Example of incorrectly resolved (WinoMT) minimal pair
(EN$\rightarrow$IT), in which both
translations realize \textit{housekeeper} as
\textit{la governante} (fem.). Translation outputs were generated by ChatGPT on July 26th, 2026.}
\label{fig_1:gender_bias_example}
\end{figure}
\vspace{-1.5em}

Over the years, several frameworks have enabled controlled gender bias evaluations across models and language pairs \citep{kocmi-etal-2020-gender,costajussa2020,choubey-etal-2021-gfst}, either by comparing translation quality across gender conditions or studying how systems resolve gender ambiguity \citep{stanovsky-etal-2019-evaluating,bentivogli-etal-2020-gender,levy-etal-2021-collecting-large}. More recently, \citet{manna-etal-2025-paying,manna-etal-2026-gender} introduced a diagnostic framework built on minimal pairs, namely otherwise identical sentences that differ only in the gender of the contextual cue (e.g., \textit{she} vs.\ \textit{he}). By testing whether models correctly adjust the corresponding gender realizations in response to the cue, this setup distinguishes context-driven disambiguation from context-independent defaults. For example, Figure~\ref{fig_1:gender_bias_example} illustrates an incorrectly resolved minimal pair, in which (Italian) translation outputs retain the same stereotypical gender realization, despite the change in the contextual cue.

Nonetheless, output-level contrasts alone cannot identify the internal components contributing to gender disambiguation or reveal how models actually utilize contextual cues when resolving gender ambiguity. In this work, we address this gap by introducing a targeted ablation framework that identifies
cue-sensitive attention heads and assesses the extent to which they support contextual gender disambiguation in state-of-the-art MT-tuned LLMs.
Our findings provide causal evidence of a strongly asymmetric mechanism: ablating a small subset of heads selectively disrupts feminine realizations, while masculine ones remain largely unaffected.

\section{Related Work}
\subsection{Gender Bias Diagnostics in MT}
Since 2019, numerous evaluation frameworks have been proposed to systematically assess gender bias in MT \citep{stanovsky-etal-2019-evaluating,bentivogli-etal-2020-gender,levy-etal-2021-collecting-large}, across language pairs and model architectures \citep{kocmi-etal-2020-gender,costajussa2020,choubey-etal-2021-gfst}. In parallel, a wide range of bias mitigation strategies have been explored, including rewriting approaches that enforce explicitly gendered or gender-neutral outputs \citep{moryossef-etal-2019-filling,habash-etal-2019-automatic,vanmassenhove-etal-2021-neutral,sun2021they, savoldi-etal-2024-prompt, piergentili-etal-2025-gender}, embedding-level debiasing techniques \citep{hirasawa-komachi-2019-debiasing,escude-font-costa-jussa-2019-equalizing}, domain adaptation \citep{saunders-byrne-2020-reducing}, and counterfactual data augmentation \citep{zmigrod-etal-2019-counterfactual}. Nonetheless, no single strategy has prevailed and the issue remains unresolved \citep{savoldi2025decade}.

This challenge persists even as LLMs increasingly underpin the state-of-the-art in MT \citep{kocmi-etal-2023-findings, kocmi-etal-2024-findings, deutsch-etal-2025-wmt24}. Despite their ability to model long-range dependencies and integrate contextual information \citep{tay2020long}, Transformer-based systems still exhibit gender bias \citep{kirk2021bias,bender2021parrots,navigli2023biases,gallegos-etal-2024-bias} and do not effectively exploit available contextual information when resolving gender ambiguity \citep{sant-etal-2024-power,vanmassenhove2024gender,manna-etal-2026-gender}. As a matter of fact, studies on intra- and inter-sentential context usage show that models often fail to effectively leverage contextual signals, either ignoring relevant information or attending to irrelevant tokens \citep{kim-etal-2019-document,goindani-shrivastava-2021-dynamic,voita-etal-2021-analyzing,yin-etal-2021-context,sarti2024_context,mohammed-niculae-2024-measuring,gu2025attentionsinkemergeslanguage}.

Considering these limitations, \citet{manna-etal-2026-gender} introduce a diagnostic framework that explicitly evaluates whether models adjust gender realizations in response to contextual gender cues (e.g., \textit{she} vs.\ \textit{he}) by jointly capturing the extent to which such cues can trigger gender disambiguation, through Minimal Pair Accuracy (MPA), as well as a model's gender assumption in the absence of explicit cues, namely its Prior Bias. Their results provide initial evidence that MT-specific post-training (e.g., instruction tuning) can improve contextual sensitivity at the output level and may influence how models process contextual gender information. However, these diagnostics alone 
cannot identify the internal mechanisms supporting such improvements or, more generally, gender disambiguation.

\subsection{Attention and Context Utilization}
Understanding whether models rely on contextual gender information (or cues) when resolving gender ambiguity requires insight into the internal mechanisms through which context is represented and integrated during inference. In Transformer-based architectures, this primarily occurs through the multi-head attention mechanism \citep{attention2017}, which enables models to dynamically combine information across tokens and layers. 

Attention weights have been widely used as a proxy for analyzing how models track token dependencies and propagate contextual information, revealing that specific attention heads may specialize in distinct linguistic functions \citep{xu2015,rocktaschel2016,wang-etal-2016-attention, lee-etal-2017-interactive,kovaleva-etal-2019-revealing, clark-etal-2019-bert, reif2019, lin-etal-2019-open, voita-etal-2019-analyzing, jo-myaeng-2020-roles}. At the same time, the interpretation of attention as explanation has been debated over the years \citep{bibal-etal-2022-attention}. As a matter of fact, high attention weights do not necessarily imply 
causal importance or proportional contribution to output logits \citep{jain-wallace-2019-attention,kobayashi-etal-2020-attention}, nor do they capture indirect information routing through intermediate heads and layers \citep{abnar-zuidema-2020-quantifying,ferrando-etal-2022-measuring,mohebbi-etal-2023-quantifying}. Nonetheless, a growing body of work has shown that attention patterns can provide useful signals related to contextual utilization \citep{olsson2022incontextlearninginductionheads,geva-etal-2023-dissecting,huangyw-etal-2025-dynamic} and has moved toward inferring the functional roles of attention heads directly from model parameters or output behavior \citep{elhelo-geva-2025-inferring,zhao-etal-2025-probing}, supporting the hypothesis that model capabilities may partially localize at the level of individual heads or small attention circuits.

While structured pruning interventions, ranging from removing entire layers \citep{fan2020reducing,sajjad2022effectdroppinglayerspretrained}, to specific attention heads \citep{michel2019sixteen,voita-etal-2019-analyzing,belhaouari2025efficient}, intermediate dimensions \citep{mccarley2021structuredpruningbertbasedquestion,wang-etal-2020-structured} or blocks in weight matrices \citep{lagunas-etal-2021-block}, have traditionally been explored for model compression and efficiency, more recent work has applied targeted head pruning to identify sub-networks responsible for specific linguistic capabilities, transfer effects, or biased behavior \citep{dasu2025attentionpruningautomatedfairness,hu-etal-2025-circuits,elhelo-geva-2025-inferring,zhao-etal-2025-probing}. Along this line, recent evidence shows that restricting post-training (or alignment) updates to task-sensitive heads can enhance or preserve model capabilities while improving efficiency \citep{chen-etal-2025-alps,zhao-etal-2025-probing}. Together, these findings indicate that attention heads provide a meaningful level of analysis for studying how contextual information is encoded and utilized in Transformer-based models, as well as a principled locus for targeted interventions.

\section{Experimental Setup}
We investigate how decoder-only LLMs that underpin the state-of-the-art in MT utilize contextual gender cues to resolve gender ambiguity, across multiple model families and (target) languages.

At the output level, we evaluate whether models adapt gender realizations to the available contextual cues and quantify the extent to which they exhibit gender preferences in ambiguous contexts. At the mechanistic level, we analyze attention patterns to identify cue-sensitive attention heads and introduce a targeted ablation setup to test their causal contribution to contextual gender disambiguation.

\subsection{Data}

Our evaluation setup is based on the diagnostic framework proposed by \citet{manna-etal-2026-gender}, which reorganizes the template-based WinoMT benchmark \citep{stanovsky-etal-2019-evaluating} 
into 1582 minimal pairs of sentences differing only in the contextual gender cue (e.g., \textit{the housekeeper \dots\ she} vs.\ \textit{he}, Figure~\ref{fig_1:gender_bias_example}), together with a neutralized condition in which the same cue is replaced with an underspecified form (e.g., \textit{they/them}). This design supports two complementary metrics: \textbf{Minimal Pair Accuracy (MPA)}, which measures whether models correctly adapt gender realizations in response to changes in the contextual cue (\textit{i.e.}, the proportion of minimal pairs for which both translations exhibit the expected gender realization), and \textbf{Prior Bias}, which captures how models distribute masculine or feminine realizations in the absence of a (binary) gender cue. The framework further distinguishes between female-dominated (Pro-F) and male-dominated (Pro-M) professions, to get insights into the interaction between contextual gender disambiguation and gender stereotypes. Additional details on the metrics are provided in Appendix~\ref{app:code}.

For the mechanistic analysis, we restrict our experiments to the subset of minimal pairs that are correctly resolved by each model under full capacity (\textit{i.e.}, before ablation), to ensure that observed effects will reflect variations in contextual (gender) integration rather than baseline translation errors.

\subsection{Models}
We consider two families of MT-tuned LLMs together with their base models.
\begin{itemize}
\item \textbf{TowerInstruct} \citep{alves2024tower}, an instruction-tuned variant of \textbf{Llama2} \citep{touvron2023b}, optimized for translation-related tasks.
\item \textbf{GemmaX2} \citep{cui-etal-2025-multilingual}
 and \textbf{TowerPlus} \citep{rei-etal-2026-tower}, which largely follow the same post-training pipeline as TowerInstruct and are built on \textbf{Gemma2} \citep{gemmateam2024gemma2}. 
\end{itemize}

Importantly, these models have been shown to outperform open-source general-purpose LLMs and achieve comparable performance to leading closed-source systems on standard MT benchmarks \citep{alves2024tower,cui-etal-2025-multilingual,rei-etal-2026-tower}, thereby representing the current state-of-the-art in MT.  

To simulate realistic MT deployment scenarios while accounting for our own computational constraints, we select the 7B and 9B variants of our Llama2- and Gemma2-based models, respectively.

\subsection{Translation Setup}
We adopt a zero-shot prompting strategy that reproduces the instruction format used during each model's instruction tuning\footnote{\url{https://huggingface.co/datasets/Unbabel/TowerBlocks-v0.2}}, and provide a natural language equivalent for each base model. In practice, this corresponds to prompts of the form: \texttt{Translate the following text from English into \{target\}}, with minor variations.

Translations are generated from English into four grammatical gender languages, namely \textbf{Italian, Spanish, French, and German}. These languages provide a controlled setting for studying gender disambiguation, as gender information is (typically) made explicit in the target even when underspecified in the source, creating the condition for which models must rely on contextual cues to correctly realize gender morphology across articles and nouns (e.g., \textit{il professore} vs.\ \textit{la professoressa} in Italian, \textit{el profesor} vs.\ \textit{la profesora} in Spanish, etc.). 

\section{Behavioral Evidence: Gender Disambiguation and Prior Bias}\label{sec:behavioural}
We first analyze the extent to which models resolve gender ambiguity by jointly evaluating \textbf{MPA} and \textbf{Prior Bias} across target languages, comparing each MT-tuned variant with its base model.  

\begin{table}[b!]
\centering
\small
\renewcommand{\arraystretch}{1.2}
\setlength{\tabcolsep}{5pt}
\resizebox{\linewidth}{!}{
\begin{tabular}{lccccc}
\toprule
& & & & \multicolumn{2}{c}{\textbf{Prior Bias}} \\
\cmidrule(lr){5-6}
\textbf{Model} & \textbf{MPA} & \textbf{Pro-F} & \textbf{Pro-M}
& Masc. & Fem. \\
\midrule
\rowcolor{lightblue}
\multicolumn{6}{c}{\textbf{en$\rightarrow$it}} \\
\midrule
Llama2 & 12.9\% & 75.0\% & 25.0\% & 85.8\% & 14.2\% \\
\rowcolor{lightgreen}
TowerInstruct & \textbf{33.3\%} & 50.5\% & 49.5\% & 76.3\% & 23.7\% \\
\addlinespace[10pt]
Gemma2 & 10.2\% & 74.1\% & 25.9\% & 85.7\% & 14.3\% \\
\rowcolor{lightgreen}
GemmaX2 & 43.2\% & 57.5\% & 42.5\% & 78.1\% & 21.9\% \\
\rowcolor{lightgreen}
TowerPlus & \textbf{64.3\%} & 44.0\% & 56.0\% & 89.1\% & 10.9\% \\

\midrule
\rowcolor{lightblue}
\multicolumn{6}{c}{\textbf{en$\rightarrow$es}} \\
\midrule
Llama2 & 16.4\% & 74.6\% & 25.4\% & 85.6\% & 14.4\% \\
\rowcolor{lightgreen}
TowerInstruct & \textbf{43.6\%} & 58.1\% & 41.9\% & 77.7\% & 22.3\% \\
\addlinespace[10pt]
Gemma2 & 20.4\% & 87.0\% & 13.0\% & 87.4\% & 12.6\% \\
\rowcolor{lightgreen}
GemmaX2 & 72.2\% & 48.4\% & 51.6\% & 73.4\% & 26.6\% \\
\rowcolor{lightgreen}
TowerPlus & \textbf{87.0\%} & 48.2\% & 51.8\% & 92.8\% & 7.2\% \\

\midrule
\rowcolor{lightblue}
\multicolumn{6}{c}{\textbf{en$\rightarrow$fr}} \\
\midrule
Llama2 & 15.5\% & 74.3\% & 25.7\% & 81.4\% & 18.6\% \\
\rowcolor{lightgreen}
TowerInstruct & \textbf{37.9\%} & 50.8\% & 49.2\% & 73.3\% & 26.7\% \\
\addlinespace[10pt]
Gemma2 & 10.4\% & 81.1\% & 18.9\% & 83.7\% & 16.3\% \\
\rowcolor{lightgreen}
GemmaX2 & 55.8\% & 45.2\% & 54.8\% & 73.0\% & 27.0\% \\
\rowcolor{lightgreen}
TowerPlus & \textbf{70.5\%} & 43.7\% & 56.3\% & 90.6\% & 9.4\% \\

\midrule
\rowcolor{lightblue}
\multicolumn{6}{c}{\textbf{en$\rightarrow$de}} \\
\midrule
Llama2 & 20.4\% & 72.4\% & 27.6\% & 87.3\% & 12.7\% \\
\rowcolor{lightgreen}
TowerInstruct & \textbf{46.5\%} & 56.6\% & 43.4\% & 77.2\% & 22.8\% \\
\addlinespace[10pt]
Gemma2 & 23.9\% & 82.0\% & 18.0\% & 86.1\% & 13.9\% \\
\rowcolor{lightgreen}
GemmaX2 & 68.1\% & 49.3\% & 50.7\% & 69.1\% & 30.9\% \\
\rowcolor{lightgreen}
TowerPlus & \textbf{79.9\%} & 48.3\% & 51.7\% & 89.4\% & 10.6\% \\

\bottomrule
\end{tabular}
}
\caption{Minimal Pair Accuracy (MPA), profession-type distribution (Pro-F, Pro-M), and Prior Bias (Masc./Fem.) across translation directions. Results are grouped by model family; MT-tuned models are highlighted.}
\label{tab:behavioural_results}
\end{table}

As reported in Table~\ref{tab:behavioural_results}, both base models exhibit relatively low MPA scores, indicating limited sensitivity to contextual gender cues. After instruction tuning, performance increases across all target languages. In particular, TowerPlus achieves the highest MPA, ranging from 64.3\% to 87.0\%. Similar improvements, relative to their respective base models, are observed for TowerInstruct and GemmaX2. Overall, these results indicate that MT-specific post-training consistently strengthens the ability of LLMs to condition gender realization in response to contextual information.

An analysis across profession types further reveals that gender disambiguation is influenced by stereotypical gender associations. Consistent with the findings of \citet{manna-etal-2026-gender}, base models show a pronounced asymmetry: correctly resolved minimal pairs predominantly involve  female-dominated professions (Pro-F), while adapting gender morphology when contextual cues require feminine realizations of male-dominated professions (Pro-M) appears more challenging. In contrast, 
MT-tuned variants exhibit a more balanced distribution, signaling improved sensitivity to contextual cues even when these conflict with stereotypical gender associations.

Nonetheless, our Prior Bias results reveal persistent masculine default preferences across all model families and target languages. 
Interestingly, TowerPlus retains the strongest masculine prior (ranging from 89.1\% to 92.8\% across languages) while also achieving the highest MPA, indicating that improved contextual gender disambiguation does not necessarily coincide with weaker gender priors. Rather, MT-specific post-training appears to strengthen the model’s ability to override prior assumptions when explicit contextual cues are available, even when masculine default preferences are retained. 

The observed behavioral patterns generalize across (target) languages, with small performance differences but a stable model ranking.

\section{Causal Evidence: Gender Encoding in Attention Heads}
Our behavioral experiments show that MT-specific post-training improves sensitivity to contextual gender cues across all model families and (target) languages. We now investigate how contextual gender information is internally represented and utilized in these models by (i) identifying cue-sensitive attention heads and (ii) testing whether these heads causally support correct gender realizations.

\begin{figure}[h!]
\centering

\includegraphics[width=\columnwidth]{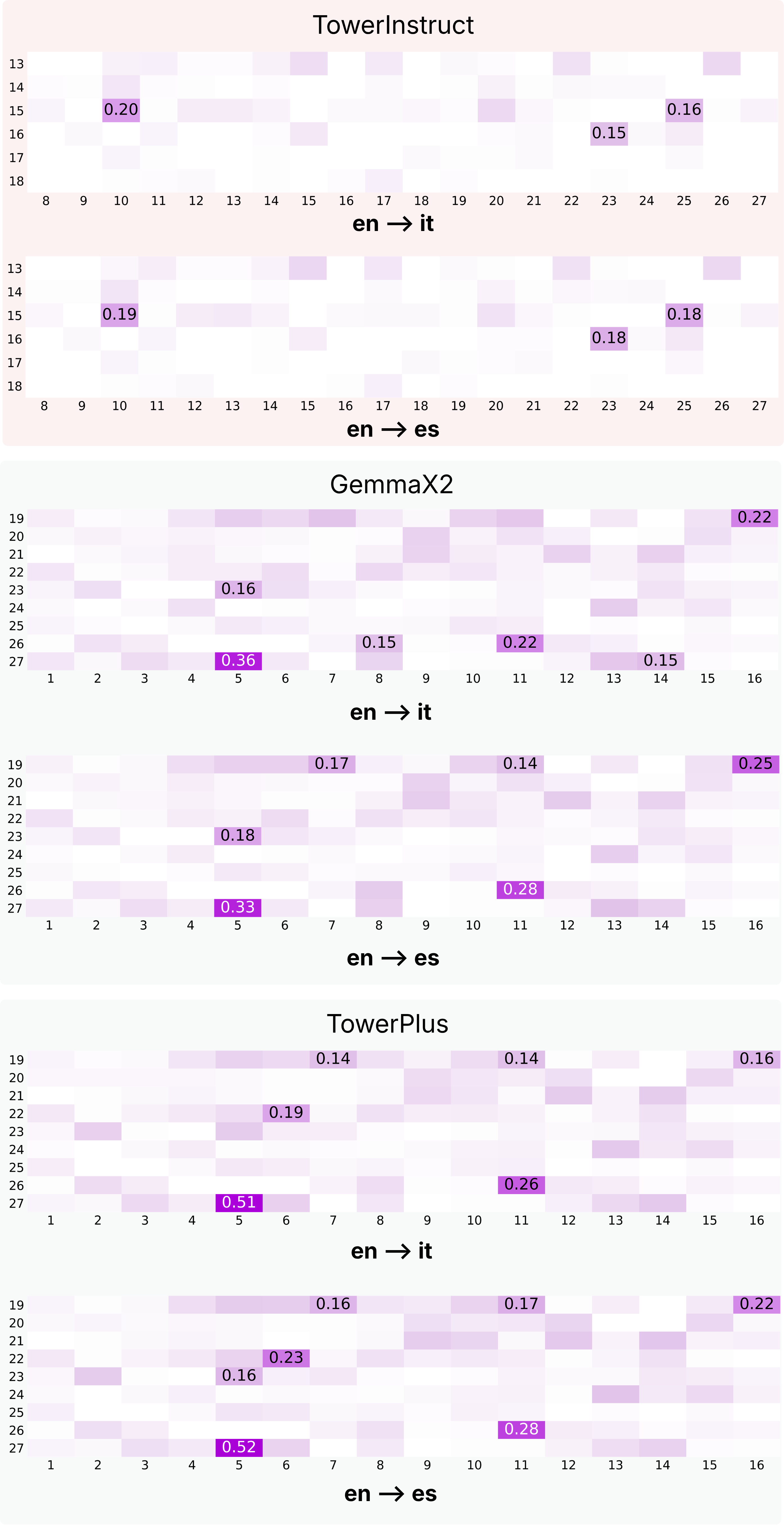}

\caption{Snapshot of attention to the \femalehl{feminine cue} over the full set of correctly resolved minimal pairs in each MT-tuned model, across two (exemplary) target languages. Layers are shown on the y-axis and heads on the x-axis; most prominent scores are annotated. Feminine contexts are selected, as they exhibit the most pronounced head-level patterns. 
}
\label{fig:att_all}
\end{figure}

\subsection{Identifying Cue-Sensitive Attention Heads}\label{sec:cue-attention}
We rank attention heads according to the \textbf{average self-attention} assigned to the gender cue (e.g., \textit{he/she}) at the timestep(s) in which the gender-marked profession is generated. Because this is typically tokenized into multiple subwords, we retrieve the attention weights over the full tokenized span, explicitly including gender-marked articles (\textit{il, lo, la} in Italian, \textit{el, la} in Spanish, etc.). These provide the first overt gender realization for the investigated grammatical gender languages. 

We then average the attention weights across subword tokens to enable the comparison across gendered forms whose tokenization may differ. The resulting scores are then aggregated over correctly resolved minimal pairs, for each MT-tuned model (TowerInstruct, GemmaX2 and TowerPlus). To account for any potential gender-specific attention asymmetries, we compute the rankings separately for masculine and feminine contexts, conditioned on the gendered pronoun. For readability, the heatmaps only display a subset of the most prominent layers and heads (Figures~\ref{fig:att_all},~\ref{fig:appendix_llama2_correct} and~\ref{fig:appendix_gemma2_correct}). However, the analysis spans \textbf{all attention heads across all layers}.

While prior work has shown that models tend to attend more strongly to feminine cues \citep{manna-etal-2025-paying,manna-etal-2026-gender}, we further observe that cue attention appears to increase after MT tuning, with the largest changes concentrated in a subset of cue-sensitive heads that is largely shared across target languages (Figure \ref{fig:att_all}). This observation is supported by significant aggregate increases across most base-to-tuned settings, with a subset of individual heads (17.3\% to 49.7\%) showing significant positive changes (Appendices~\ref{app:attention_gender_comparison},~\ref{app:attention_tuning}) 

Nonetheless, these patterns do not by themselves establish that the identified heads causally contribute to gender disambiguation. Rather, they provide the criterion used to select candidate heads as targets of our subsequent ablation analysis.

\begin{figure*}[h!]
\centering
\includegraphics[width=\linewidth]{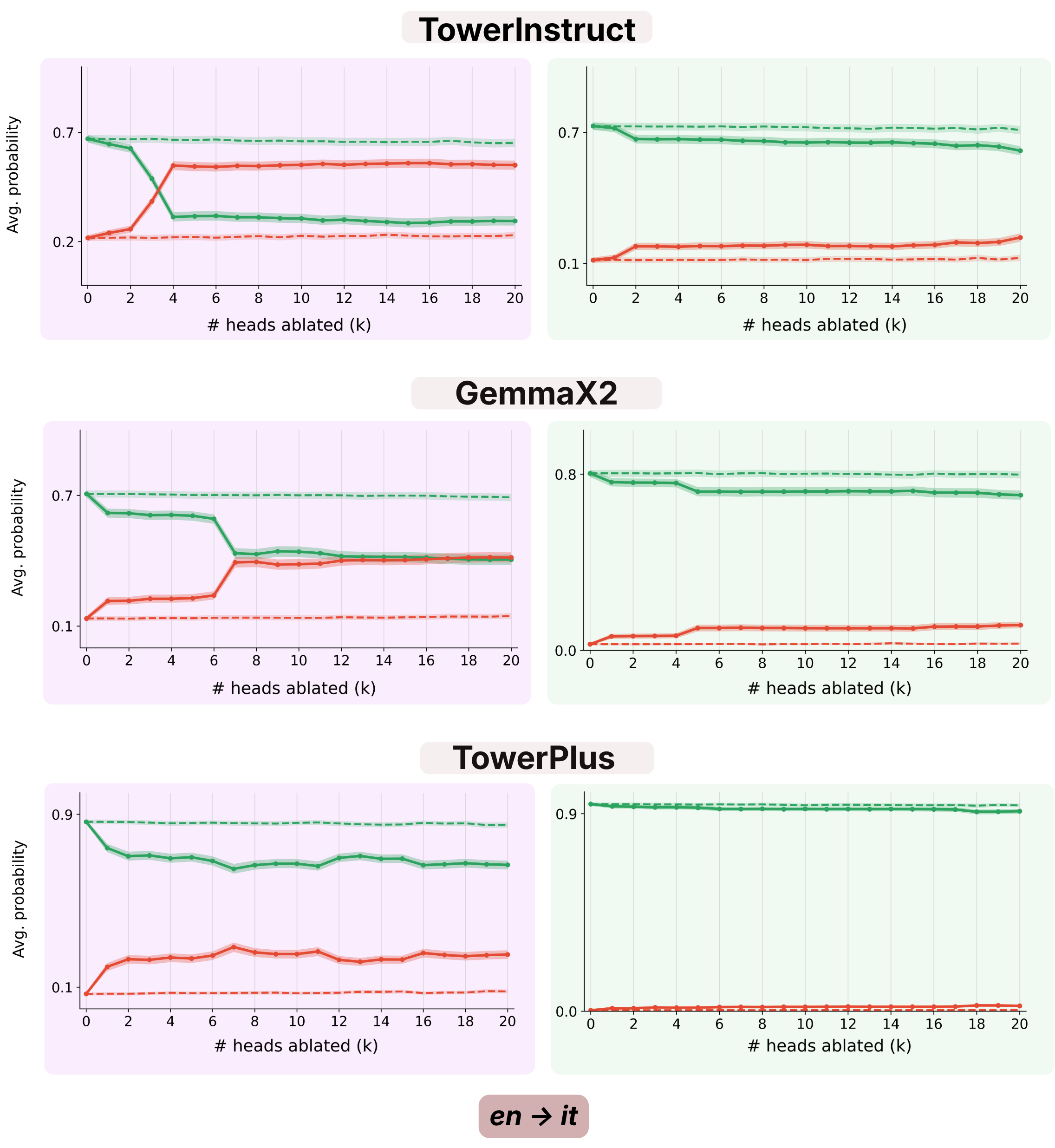}
\caption{Ablation effects for en$\rightarrow$it across models. Panels correspond to \femalehl{feminine} and \malehl{masculine} contexts. Solid curves show the probability assigned to the correct (green) and incorrect (red) gender; dashed curves show random-ablation controls. Shaded areas denote 95\% CI across examples. Sample sizes (post filtering) are
$n=358$ for TowerInstruct, $n=403$ for GemmaX2, and $n=673$ for
TowerPlus.}\label{fig:ablation}
\end{figure*}

\subsection{Progressive Head Ablation}
To determine whether the identified cue-sensitive heads play a causal role in gender disambiguation, we perform a controlled mechanistic intervention. Specifically, we progressively ablate the \textbf{top-$\mathbf{k}$ ranked heads} (up to $k=20$) by jointly zeroing their outputs during each forward pass. 

We focus on the timestep at which the \textbf{gender-marked article} is generated, corresponding to the first position where grammatical gender is overtly realized across all target languages (e.g., \textit{il/lo} vs.\ \textit{la} in Italian), and therefore the point at which the model effectively commits to a gender. We then extract the probability assigned to masculine and feminine article forms, selecting the highest-probability candidate within each gender category. 

Because such a setup requires comparing unambiguous, single-token article forms, we perform the intervention on \textbf{Italian, French, and Spanish} (excluding German, whose article forms are ambiguous with respect to gender across grammatical cases). While all Italian article forms satisfy these requirements, the French and Spanish analyses are restricted to examples whose gender-marked articles are tokenized as a single token (\textit{i.e.}, \textit{le vs.\ la} and \textit{el vs.\ la}, respectively), ensuring that all candidates are drawn from the same softmax distribution and therefore provide directly comparable measures of the model's gender preference at each ablation step. Importantly, this choice is validated by an observed near-perfect agreement between the grammatical gender expressed in the (preceding) article and the profession noun across translation outputs, with agreement scores ranging between 97\% and 100\% (see Appendix~\ref{app:articles}). Moreover, this setup allows us to retain epicene profession nouns whose preceding article provides an unambiguous gender marker (e.g., \textit{le/la journaliste} in French, \textit{el/la periodista} in Spanish), while fully epicene noun phrases are excluded (e.g., \textit{l'artiste}, \textit{l'analyste} in French, etc.).

As a control condition, we repeat the same procedure by ablating \textbf{$\mathbf{k}$ randomly selected heads}. This ensures that any observed effects are specifically attributable to the removal of cue-sensitive components rather than to general model degradation. 

We perform the intervention on correctly resolved minimal pairs only, further restricting the analysis to cases in which the model assigns a higher probability to the correct gender realization under full model capacity, thereby reducing potential noise from incorrect (automatic) labels. 

\vspace{1em}
\subsection{Asymmetric Effects on Gender Disambiguation}

For each target language (Italian, Spanish, French), we track the average probability assigned to correct and competing gender realizations as cue-sensitive heads are progressively ablated.

\begin{figure}[h!]
\vspace{1em}
\centering
\includegraphics[width=\linewidth]{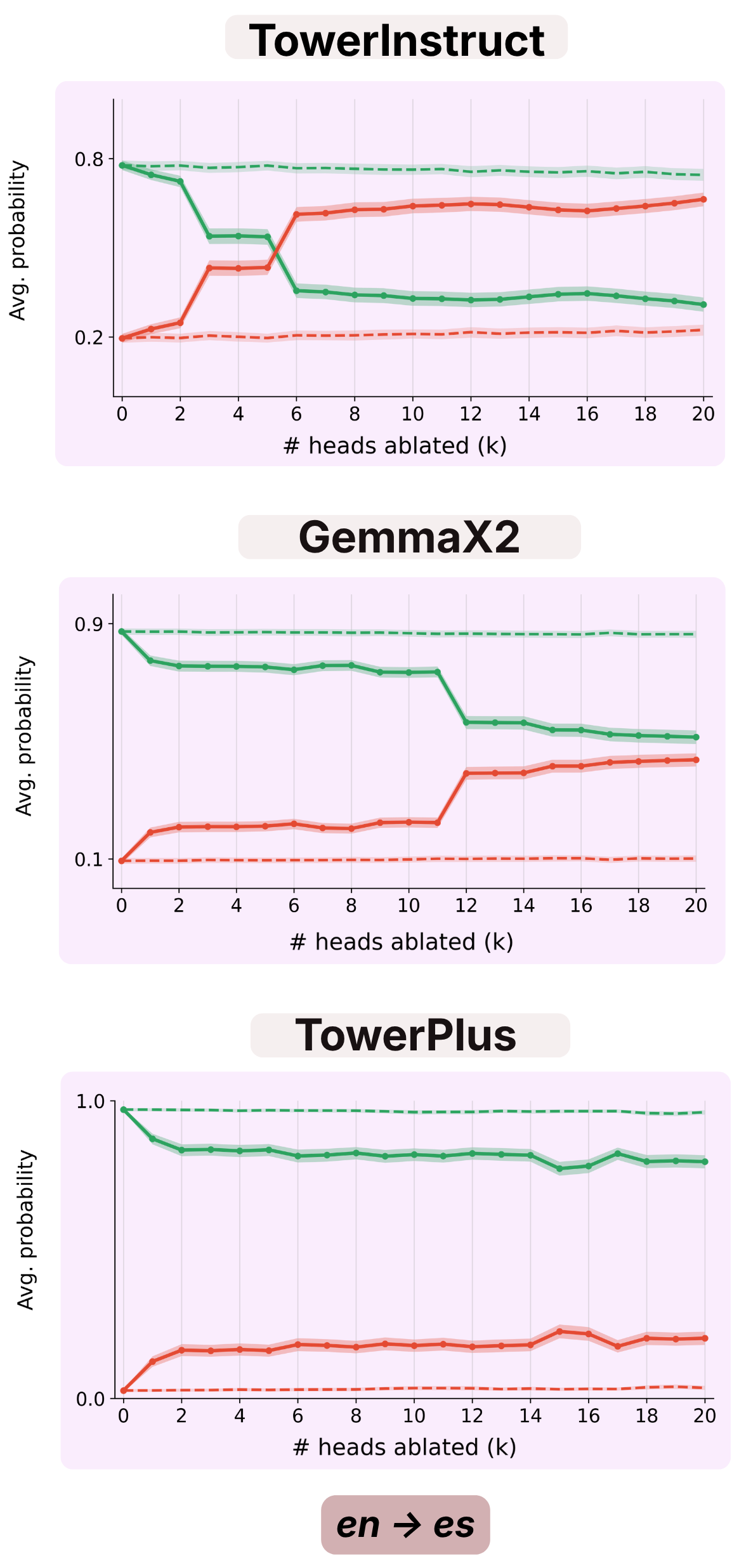}
\caption{Ablation effects for en$\rightarrow$es across models,
shown for \femalehl{feminine} contexts only. Solid curves show the probability assigned to the correct (green) and incorrect (red) gender realization; dashed curves show random-ablation controls. Shaded areas denote 95\% CI across examples. Sample sizes (post filtering, minimal pairs) are $n=336$ for TowerInstruct, $n=521$ for GemmaX2, and $n=656$ for TowerPlus.}\label{fig:ablation2}
\end{figure}
The effect is strongly asymmetric: in feminine contexts, ablating a small number of cue-sensitive heads leads to substantial (probability mass) shifts toward incorrect (masculine) realizations, whereas masculine contexts remain effectively invariant. Figure~\ref{fig:ablation} illustrates this asymmetry using Italian as a representative case, while the ablation curves for all model-language settings are provided in
Appendix~\ref{app:full_ablation_curves}.

The magnitude of the ablation effect nevertheless varies across models and languages. In TowerInstruct, the probability of the correct realization is rapidly overtaken by the competing (masculine) alternative after ablating only a few heads in all three languages ($k=4$ for Italian, $k=5$ for French, $k=6$ for Spanish). GemmaX2 shows a first drop as the top-ranked head is ablated, followed by a second, larger one (at $k=7$ for Italian and French, but only at $k=12$ for Spanish, see Figure~\ref{fig:ablation2}) before the two probabilities slowly converge. TowerPlus, instead, shows a more gradual decline across all three languages, which hints at more robust or diffuse gender encoding mechanisms.

Crucially, these shifts are absent under random ablation: probabilities remain largely unchanged when up to 20 heads are removed, indicating that the observed degradation is specifically attributable to the identified cue-sensitive heads rather than general model disruption.

\begin{table}[h!]
\centering
\small
\setlength{\tabcolsep}{8pt}
\renewcommand{\arraystretch}{1.05}
\resizebox{\columnwidth}{!}{
\begin{tabular}{l ccc >{\columncolor{fem}}c >{\columncolor{masc}}c}
\toprule
\textbf{Model} & $\mathbf{n}$ & \textbf{\# heads ($\mathbf{k}$)} & \textbf{\% heads} &
\multicolumn{2}{c}{\textbf{Flip Rate}} \\
\midrule
\rowcolor{lightblue}
\multicolumn{6}{c}{\textbf{en$\rightarrow$it}} \\
\midrule
TowerInstruct & 358 & 4  & 0.4\% & 74.3\% & 9.5\% \\
GemmaX2       & 403 & 12 & 1.8\% & 49.6\% & 8.2\% \\
TowerPlus     & 673 & 7  & 1.0\% & 27.2\% & 0.6\% \\

\midrule
\rowcolor{lightblue}
\multicolumn{6}{c}{\textbf{en$\rightarrow$fr}} \\
\midrule
TowerInstruct & 398 & 5  & 0.5\% & 79.1\% & 4.3\% \\
GemmaX2       & 603 & 17 & 2.5\% & 51.9\% & 8.1\% \\
TowerPlus     & 765 & 16 & 2.4\% & 27.8\% & 1.4\% \\

\midrule
\rowcolor{lightblue}
\multicolumn{6}{c}{\textbf{en$\rightarrow$es}} \\
\midrule
TowerInstruct & 336 & 6  & 0.6\% & 74.7\% & 6.2\% \\
GemmaX2       & 521 & 17 & 2.5\% & 51.1\% & 5.0\% \\
TowerPlus     & 656 & 15 & 2.2\% & 19.2\% & 0.6\% \\

\bottomrule
\end{tabular}
}
\caption{Flip rates after ablating the top-$k$ cue-sensitive heads in \femalehl{feminine} and \malehl{masculine} contexts. For each model and (target) language, $k$ denotes the point after which the ablation curves begin to stabilize; \% heads is $k$ as a share of all attention heads (1024 for TowerInstruct, 672 for GemmaX2 and TowerPlus).}
\label{tab:flip_rates}
\end{table}

While these curves summarize average effects, they do not directly quantify how often the model's preferred realization actually changes. We therefore compute the \textbf{\textit{flip rate}}, namely the proportion of cases in which the full model prefers the correct gender but switches to the competing realization after ablation. As reported in Table~\ref{tab:flip_rates}, the flip rates closely mirror the probability curves: TowerInstruct consistently exhibits the largest feminine flip rates, TowerPlus the smallest. Masculine flip rates, by contrast, remain
very low across all models and languages. 
Moreover, we find that anti-stereotypical gender realizations generally show higher flip rates than pro-stereotypical ones across both gender conditions (Appendix~\ref{app:flip_stereotype}).

Overall, our results provide direct evidence that a small subset of the identified cue-sensitive heads is causally involved in the encoding and integration of contextual gender information in our models' decision process. As a matter of fact, ablating only a small number of cue-sensitive heads ($< 3\%$ of all attention heads) not only reduces confidence in the (correct) feminine realization, but frequently causes the model to switch to the masculine alternative altogether. Masculine predictions, by contrast, remain largely stable across all evaluated models and target languages.

\section{Discussion}
While prior work has provided initial evidence that MT-specific post-training can improve contextual sensitivity at the output level \citep{manna-etal-2026-gender}, we find that these improvements are consistent across model families and target languages, yet do not necessarily weaken a model's gender priors. Instead, enhanced contextual sensitivity appears to operate as a conditional correction mechanism rather than through a symmetric encoding of gender information. Models seem to require explicit contextual grounding to correctly produce feminine realizations and override a default-to-masculine reasoning process, consistently documented across language models and MT systems \citep{jumelet-etal-2019-analysing,van-der-wal-etal-2022-birth,hackenbuchner-etal-2025-genderous, klein2026relevantbertignoresexplicit}. 

Our structured attention-head ablation experiments provide causal evidence for this interpretation. Across models and languages, removing less than 3\% of all attention heads selectively disrupts feminine realizations, frequently shifting predictions toward the competing masculine form, while masculine realizations remain largely stable. A comparable cross-modal asymmetry has been observed in speech translation, where \citet{conti-etal-2025-unheard} find that removing gender-relevant acoustic information produces substantially more feminine-to-masculine than masculine-to-feminine flips.

Moreover, our findings contribute to the ongoing debate on the relevance of attention in interpretability research \citep{jain-wallace-2019-attention, bibal-etal-2022-attention}. While raw attention weights alone only provide correlational insights, our mechanistic analysis supports the view that model capabilities (or biases) may, at least partially, localize at the level of individual heads or small attention circuits \citep{dasu2025attentionpruningautomatedfairness,hu-etal-2025-circuits,elhelo-geva-2025-inferring,zhao-etal-2025-probing}. Nonetheless, the identified heads should not be understood as the only pathway through which gender information is propagated within the models. Because candidate heads are selected and ranked based on direct attention to the cue, indirect information routing (e.g., through intermediate heads and token positions) is not captured \citep{abnar-zuidema-2020-quantifying,ferrando-etal-2022-measuring,mohebbi-etal-2023-quantifying}. Similarly, we cannot rule out that masculine signals may be mediated by a separate or more distributed mechanism involving other model components.

Beyond interpretability, identifying restricted attention circuits that are responsible for the encoding of gender information provides a principled direction for targeted bias mitigation strategies. While controlling gender distributions in large-scale \mbox{(pre-)training} data remains a challenge, focusing on the adaptation or alignment of relevant model components may offer a more scalable approach to making MT models fairer.

\section{Conclusion}
We investigate how MT-tuned LLMs resolve gender ambiguity by combining behavioral and causal evidence. Across all models and (target) languages, MT-specific post-training consistently enhances contextual sensitivity without necessarily weakening underlying gender priors. Instead, it appears to reorganize the balance between contextual evidence and prior gender preferences, enabling models to override prior
gender assumptions when explicit (feminine) cues are available.

We provide causal evidence that this process relies, at least in part, on a small subset of cue-sensitive attention heads, whose ablation disproportionately disrupts feminine realizations, while masculine predictions remain comparatively stable. More broadly, we establish a causal link between attention and context utilization, validating attention heads as a meaningful target for understanding and potentially correcting biased behavior.

\section*{Limitations}
We acknowledge that our analysis is restricted to non-ambiguous contexts in which gender information is explicitly available in the source and overtly realized in the target. Both levels of analysis rely on frameworks that encode gender as a binary variable, thereby excluding non-binary and gender-neutral alternatives, for which a growing number of resources are emerging across languages \citep{friidhriksdottir-2024-genderqueer,savoldi-etal-2025-mind,jourdan2025fairtranslate,pranav-etal-2025-glitter,chen2024ambgimt}. For instance, fully epicene (\textit{i.e.}, gender neutral) realizations, in which neither the article nor the profession noun provides an overt gender contrast, are excluded from the analysis. 
While this setup enables a controlled investigation of contextual gender disambiguation in MT, particularly considering that state-of-the-art systems still struggle with neutral or non-binary gender translations \citep{hackenbuchner-etal-2025-genderous,pranav-etal-2025-glitter}, it is necessarily simplified. 


Moreover, although we observe remarkably consistent output-level trends and ablation effects across target languages, our analysis is limited to a set of closely related grammatical gender languages. As a result, it remains unclear to what extent the same components would emerge in typologically different languages or in settings with substantially distinct morphological realizations of gender. 

Finally, our framework is primarily diagnostic: while we identify and causally validate a small subset of cue-sensitive attention heads involved in contextual gender disambiguation, we do not directly intervene on these mechanisms to improve or debias model behavior. A promising direction for future work is to make these findings actionable, for instance by turning the identified circuits into a locus for targeted interventions.

\bibliography{custom,anthology-1,anthology-2}
\appendix

\section{Evaluation Procedure}\label{app:code}
Our evaluation builds upon the diagnostic framework introduced by \citet{manna-etal-2026-gender}, itself based on WinoMT \citep{stanovsky-etal-2019-evaluating}. The benchmark is reorganized into 1582 minimal pairs that differ only in the contextual gender cue (e.g., \textit{the housekeeper \dots\ he} vs.\ \textit{the housekeeper \dots\ she}); each pair is complemented by a neutralized condition in which the gendered cue is replaced by an underspecified form (e.g., \textit{they/them}). The evaluation pipeline relies on \texttt{fast\_align} \citep{dyer-etal-2013-simple} for word alignment (Apache License 2.0) and \texttt{spaCy}'s morphological analyzer \citep{honnibal2020spacy} (MIT license) to extract the grammatical gender of the aligned target-side profession noun. These annotations are then used to compute the following metrics:

\paragraph{\textbf{Minimal Pair Accuracy (MPA)}} measures the proportion of minimal pairs for which both translations exhibit the expected gender realization following the contextual cue. For example, an English pair containing \textit{housekeeper \dots\ he} and \textit{housekeeper \dots\ she} is counted as correct when the corresponding translations realize the expected masculine and feminine forms, respectively (e.g., Italian \textit{il governante} vs.\ \textit{la governante}).

\paragraph{\textbf{Prior Bias}} is computed from the neutralized condition and captures the distribution of masculine and feminine realizations produced when no binary gender cue is available in the source (e.g., \textit{housekeeper \dots\ they} $\rightarrow$ \textit{il governante} vs.\ \textit{la governante}). 

\section{Computational Infrastructure and Reproducibility}
Our translation experiments were conducted on a single GPU, while the ablation analysis was performed on two NVIDIA L40S GPUs (46GB VRAM). 

To ensure reproducibility, all translation outputs were generated deterministically (without sampling). The ablation analysis further enforced deterministic behaviour by using greedy decoding, yielding fixed outputs for the same prompts and model parameters.

\section{Gender-Specific Differences in Attention}
\label{app:attention_gender_comparison}

Figures~\ref{fig:appendix_llama2_correct} and~\ref{fig:appendix_gemma2_correct} provide snapshots of the (cue) attention assigned to feminine and masculine cues in Llama2- and Gemma2-based models, respectively (as described in Section~\ref{sec:cue-attention}). Consistent with prior work \citep{manna-etal-2026-gender}, we can observe that feminine cues generally receive higher attention. To establish whether the observed difference is systematic, we complement our analysis with a paired statistical comparison.

\begin{table}[h!]
\centering
\fontsize{5.5}{6.5}\selectfont
\renewcommand{\arraystretch}{0.82}
\resizebox{\linewidth}{!}{
\begin{tabular}{lcccc}
\toprule
\textbf{Model} & \textbf{$n$} & \textbf{Fem.} & \textbf{Masc.} & \textbf{F $>$ M ($r_{\mathrm{rb}}$)} \\
\midrule
\rowcolor{lightblue}
\multicolumn{5}{c}{\textbf{en$\rightarrow$it}} \\
\midrule
Llama2 & \multirow{2}{*}{94} & .005 & .004 & .708$^{***}$ \\
\cellcolor{lightgreen}TowerInstruct & & \cellcolor{lightgreen}.006 & \cellcolor{lightgreen}.004 & \cellcolor{lightgreen}.953$^{***}$ \\
\addlinespace[6pt]
Gemma2 & \multirow{2}{*}{106} & .014 & .011 & .708$^{***}$ \\
\cellcolor{lightgreen}GemmaX2 & & \cellcolor{lightgreen}.015 & \cellcolor{lightgreen}.013 & \cellcolor{lightgreen}.566$^{***}$ \\
\addlinespace[6pt]
Gemma2 & \multirow{2}{*}{109} & .014 & .011 & .705$^{***}$ \\
\cellcolor{lightgreen}TowerPlus & & \cellcolor{lightgreen}.015 & \cellcolor{lightgreen}.012 & \cellcolor{lightgreen}.915$^{***}$ \\
\midrule
\rowcolor{lightblue}
\multicolumn{5}{c}{\textbf{en$\rightarrow$es}} \\
\midrule
Llama2 & \multirow{2}{*}{172} & .006 & .004 & .919$^{***}$ \\
\cellcolor{lightgreen}TowerInstruct & & \cellcolor{lightgreen}.006 & \cellcolor{lightgreen}.004 & \cellcolor{lightgreen}.937$^{***}$ \\
\addlinespace[6pt]
Gemma2 & \multirow{2}{*}{251} & .018 & .013 & .790$^{***}$ \\
\cellcolor{lightgreen}GemmaX2 & & \cellcolor{lightgreen}.017 & \cellcolor{lightgreen}.014 & \cellcolor{lightgreen}.715$^{***}$ \\
\addlinespace[6pt]
Gemma2 & \multirow{2}{*}{304} & .017 & .012 & .799$^{***}$ \\
\cellcolor{lightgreen}TowerPlus & & \cellcolor{lightgreen}.017 & \cellcolor{lightgreen}.014 & \cellcolor{lightgreen}.934$^{***}$ \\
\midrule
\rowcolor{lightblue}
\multicolumn{5}{c}{\textbf{en$\rightarrow$fr}} \\
\midrule
Llama2 & \multirow{2}{*}{132} & .005 & .004 & .853$^{***}$ \\
\cellcolor{lightgreen}TowerInstruct & & \cellcolor{lightgreen}.006 & \cellcolor{lightgreen}.004 & \cellcolor{lightgreen}.965$^{***}$ \\
\addlinespace[6pt]
Gemma2 & \multirow{2}{*}{116} & .015 & .012 & .716$^{***}$ \\
\cellcolor{lightgreen}GemmaX2 & & \cellcolor{lightgreen}.016 & \cellcolor{lightgreen}.014 & \cellcolor{lightgreen}.398$^{***}$ \\
\addlinespace[6pt]
Gemma2 & \multirow{2}{*}{138} & .015 & .011 & .746$^{***}$ \\
\cellcolor{lightgreen}TowerPlus & & \cellcolor{lightgreen}.016 & \cellcolor{lightgreen}.012 & \cellcolor{lightgreen}.852$^{***}$ \\
\midrule
\rowcolor{lightblue}
\multicolumn{5}{c}{\textbf{en$\rightarrow$de}} \\
\midrule
Llama2 & \multirow{2}{*}{197} & .006 & .005 & .837$^{***}$ \\
\cellcolor{lightgreen}TowerInstruct & & \cellcolor{lightgreen}.007 & \cellcolor{lightgreen}.004 & \cellcolor{lightgreen}.950$^{***}$ \\
\addlinespace[6pt]
Gemma2 & \multirow{2}{*}{307} & .017 & .013 & .838$^{***}$ \\
\cellcolor{lightgreen}GemmaX2 & & \cellcolor{lightgreen}.017 & \cellcolor{lightgreen}.015 & \cellcolor{lightgreen}.641$^{***}$ \\
\addlinespace[6pt]
Gemma2 & \multirow{2}{*}{326} & .017 & .012 & .840$^{***}$ \\
\cellcolor{lightgreen}TowerPlus & & \cellcolor{lightgreen}.017 & \cellcolor{lightgreen}.013 & \cellcolor{lightgreen}.937$^{***}$ \\
\bottomrule
\end{tabular}
}
\caption{Aggregate cue attention assigned for feminine and masculine contexts, averaged across layers and heads, over $n$ correctly resolved pairs. Conditions are compared using one-sided paired Wilcoxon signed-rank tests ($A_i^{F} > A_i^{M}$); with $r_{\mathrm{rb}}$ denoting the rank-biserial correlation ($^{***}p<.001$).}
\label{tab:attention_gender_comparison}
\end{table}

For each minimal pair $i$, we average cue attention across all layers and heads into one overall (model-wide) score for the feminine ($A_i^{F}$) and one for the masculine context ($A_i^{M}$) separately, and compute
\begin{equation}
d_i = A_i^{F} - A_i^{M},
\label{eq:fem_masc_attention_difference}
\end{equation}

We restrict the analysis to the common subset of minimal pairs for which both gender realizations are correctly resolved by both the base and MT-tuned model. For each model separately, we test whether the within-pair difference ($d_i$) is systematically positive using a one-sided paired Wilcoxon signed-rank test ($A_i^{F} > A_i^{M}$) and report the matched-pairs rank-biserial correlation $r_{\mathrm{rb}}$ as an effect-size measure.

This model-wide aggregation provides an overall comparison without selecting individual heads based on qualitative inspection or assuming that the same heads are responsive to feminine and masculine cues. The resulting attention values are therefore small and not directly informative. Nevertheless, the consistently positive and generally large $r_{\mathrm{rb}}$ values indicate a strong and systematic feminine advantage across minimal pairs. This difference is statistically significant across all model-language settings, in both base and MT-tuned models (all $p<0.001$; Table~\ref{tab:attention_gender_comparison}).

\section{Effect of MT-tuning on Attention}
\label{app:attention_tuning}

Figures~\ref{fig:appendix_llama2_correct} and~\ref{fig:appendix_gemma2_correct} further allow us to compare each MT-tuned variant with its base model. Across all settings, we observe that several prominent heads generally exhibit higher attention scores after MT-tuning and are largely shared across (target) languages. In the Llama2-based models, this pattern is most apparent for feminine cues, with heads such as \texttt{L15H10}, \texttt{L15H25}, and \texttt{L16H23} becoming particularly prominent. In the Gemma2-based models, increases are visible for both gender conditions, although the highest scores are generally observed in feminine contexts; prominent heads include \texttt{L19H16}, \texttt{L26H11} and \texttt{L27H5}. To determine whether these differences are systematic, we complement the analysis with both aggregate and head-level statistical comparisons.

\begin{table}[h!]
\centering
\small
\renewcommand{\arraystretch}{1.15}
\setlength{\tabcolsep}{6pt}
\resizebox{\linewidth}{!}{
\begin{tabular}{lr >{\columncolor{fem}}c@{\hspace{15pt}} >{\columncolor{masc}}c}
\toprule
& \textbf{$n$} & \multicolumn{2}{c}{\textbf{MT $>$ Base ($r_{\mathrm{rb}}$)}} \\
\midrule
\rowcolor{lightblue}
\multicolumn{4}{c}{\textbf{Llama2 $\rightarrow$ TowerInstruct}} \\
\midrule
en--it & 94  & $0.80^{***}$ & $-0.31^{\mathrm{n.s.}}$ \\
en--es & 172 & $0.37^{***}$ & $-0.46^{\mathrm{n.s.}}$ \\
en--fr & 132 & $0.42^{***}$ & $-0.38^{\mathrm{n.s.}}$ \\
en--de & 197 & $0.58^{***}$ & $-0.29^{\mathrm{n.s.}}$ \\
\midrule
\rowcolor{lightblue}
\multicolumn{4}{c}{\textbf{Gemma2 $\rightarrow$ GemmaX2}} \\
\midrule
en--it & 106 & $0.63^{***}$            & $0.81^{***}$ \\
en--es & 251 & $-0.13^{\mathrm{n.s.}}$ & $0.34^{***}$ \\
en--fr & 116 & $0.09^{\mathrm{n.s.}}$  & $0.78^{***}$ \\
en--de & 307 & $0.18^{**}$             & $0.73^{***}$ \\
\midrule
\rowcolor{lightblue}
\multicolumn{4}{c}{\textbf{Gemma2 $\rightarrow$ TowerPlus}} \\
\midrule
en--it & 109 & $0.52^{***}$ & $0.48^{***}$ \\
en--es & 304 & $0.17^{**}$  & $0.54^{***}$ \\
en--fr & 138 & $0.32^{***}$ & $0.65^{***}$ \\
en--de & 326 & $0.13^{*}$   & $0.50^{***}$ \\
\bottomrule
\end{tabular}
}
\caption{
Aggregate attention change after MT tuning, over $n$ resolved pairs, in \femalehl{feminine} and \malehl{masculine} contexts. Values report $r_{\mathrm{rb}}$ from a one-sided paired Wilcoxon test (MT $>$ Base); positive values indicate higher (aggregate) cue attention after MT tuning ($^{*}p<.05$, $^{**}p<.01$, $^{***}p<.001$).
}
\label{tab:attention_tuning_aggregate}
\end{table}

\paragraph{Aggregate analysis.}
For each minimal pair $i$, we first average the attention scores across layers and heads, for feminine and masculine contexts separately (as in Appendix~\ref{app:attention_gender_comparison}). We restrict the analysis to the shared set of minimal pairs correctly resolved by both the base and corresponding MT-tuned model. 

For each gender condition ($G \in \{F, M\}$), we compute the within-pair tuning effect
\begin{equation}
d_i^G =
A_{i,\mathrm{MT}}^G - A_{i,\mathrm{Base}}^G,
\label{eq:agg_delta}
\end{equation}

We test whether $d_i^G$ is systematically positive using a one-sided Wilcoxon signed-rank test ($A_i^{MT}$>$A_i^{base}$), and report the matched-pairs rank-biserial correlation ($r_{\mathrm{rb}}$). Positive values indicate higher aggregate cue attention after MT tuning.

\paragraph{Head-level analysis.}
At the individual head level $h$, we compute the mean base-to-tuned change:

\begin{equation}
d_h^G =
\frac{1}{N}\sum_{i=1}^{N}
\left(
A_{i,h,\mathrm{MT}}^G - A_{i,h,\mathrm{Base}}^G
\right).
\label{eq:head_delta}
\end{equation}

Each head is tested over paired items using a one-sided paired Wilcoxon signed-rank test ($A_{i,h,\mathrm{MT}}^G > A_{i,h,\mathrm{Base}}^G$).
We then report the percentage of heads showing a significant positive change, together with the largest mean head-level increase, $\max\Delta_h^G$, for masculine and feminine contexts separately.

\begin{table}[h!]
\centering
\small
\renewcommand{\arraystretch}{1.15}
\setlength{\tabcolsep}{6pt}
\resizebox{\linewidth}{!}{
\begin{tabular}{lr@{\hspace{25pt}} >{\columncolor{fem}}c >{\columncolor{masc}}c @{\hspace{15pt}}
                    >{\columncolor{fem}}c >{\columncolor{masc}}c}
\toprule
& \textbf{$n$} & \multicolumn{2}{c}{\textbf{$\max\Delta_h$}} & \multicolumn{2}{c}{\textbf{heads $\uparrow$ (\%)}} \\
\midrule
\rowcolor{lightblue}
\multicolumn{6}{c}{\textbf{Llama2 $\rightarrow$ TowerInstruct}} \\
\midrule
en--it & 94  & 0.12 & 0.07 & 25.3 & 17.3 \\
en--es & 172 & 0.15 & 0.07 & 21.9 & 19.5 \\
en--fr & 132 & 0.11 & 0.05 & 23.3 & 19.8 \\
en--de & 197 & 0.13 & 0.06 & 26.5 & 21.3 \\
\midrule
\rowcolor{lightblue}
\multicolumn{6}{c}{\textbf{Gemma2 $\rightarrow$ GemmaX2}} \\
\midrule
en--it & 106 & 0.29 & 0.28 & 42.7 & 45.2 \\
en--es & 251 & 0.26 & 0.30 & 43.5 & 45.4 \\
en--fr & 116 & 0.26 & 0.27 & 38.1 & 44.9 \\
en--de & 307 & 0.34 & 0.30 & 43.3 & 49.0 \\
\midrule
\rowcolor{lightblue}
\multicolumn{6}{c}{\textbf{Gemma2 $\rightarrow$ TowerPlus}} \\
\midrule
en--it & 109 & 0.42 & 0.31 & 47.8 & 46.1 \\
en--es & 304 & 0.43 & 0.35 & 48.7 & 47.6 \\
en--fr & 138 & 0.41 & 0.32 & 47.8 & 46.0 \\
en--de & 326 & 0.46 & 0.34 & 48.8 & 49.7 \\
\bottomrule
\end{tabular}
}
\caption{
Head-level attention changes after MT tuning, over $n$ resolved pairs, in \femalehl{feminine} and \malehl{masculine} contexts. $\max\Delta_h$ is the largest per-head mean change; ``heads $\uparrow$ (\%)'' is the percentage of heads with a significant positive change (one-sided paired Wilcoxon, Benjamini-Hochberg corrected).
}
\label{tab:attention_tuning_headlevel}
\end{table}

\paragraph{Results.}
In most settings, attention is significantly higher after MT tuning, with some model-specific variations (Table~\ref{tab:attention_tuning_aggregate}). Specifically, TowerInstruct shows significant increases for feminine cues across all four (target) languages, but no significant increase for masculine cues. GemmaX2 exhibits a more heterogeneous pattern: increases are significant for both contexts in Italian and German, and only for masculine ones in Spanish and French. Finally, TowerPlus (the best performing model) shows the most consistent effect, with significant increases for both feminine and masculine contexts across all (target) languages. Nevertheless, averaging attention changes across the entire network may dilute localized effects concentrated in a small number of heads.

As a matter of fact, the head-level results show that the changes are not uniform across the network (Table~\ref{tab:attention_tuning_headlevel}). The percentage of heads showing a significant positive change ranges from 17.3\% to 26.5\% for TowerInstruct, from 38.1\% to 49.0\% for GemmaX2 and from 46.0\% to 49.7\% for TowerPlus, with the latter also exhibiting the largest individual head-level increases ($\max\Delta_h$). 





\section{Full Ablation Curves}
\label{app:full_ablation_curves}

\begin{figure*}[h!]
\centering
\includegraphics[width=\linewidth]{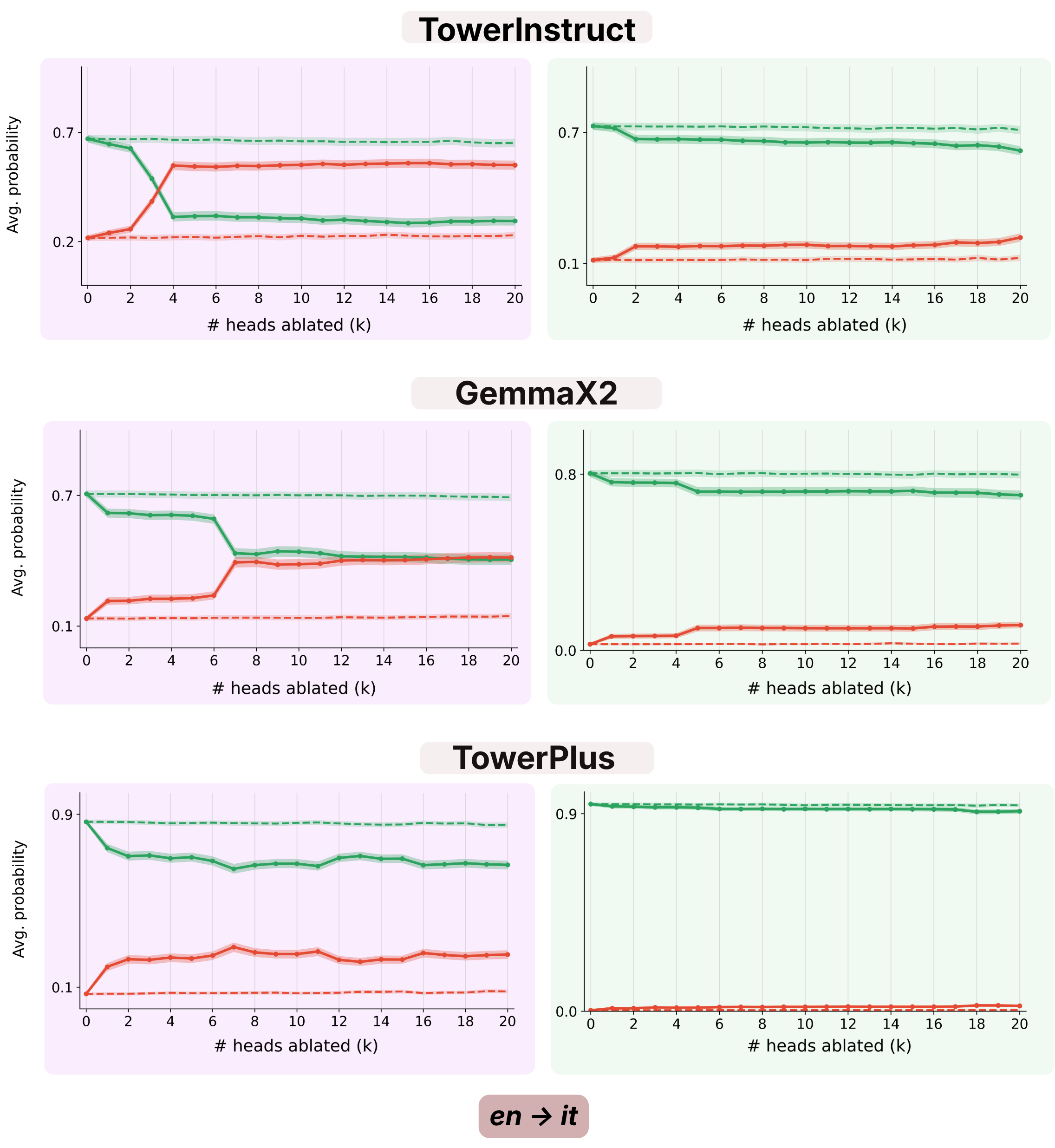}

\caption{Ablation effects for en$\rightarrow$it across models. Panels correspond to \femalehl{feminine} and \malehl{masculine} contexts. Solid curves show the probability assigned to the correct (green) and incorrect (red) gender; dashed curves show random-ablation controls. Shaded areas denote 95\% CI across examples. Sample sizes (post filtering) are $n=358$ for TowerInstruct, $n=403$ for GemmaX2, and $n=673$ for TowerPlus.}
\label{fig:full_ablation_it}
\end{figure*}

\begin{figure*}[h!]
\centering
\includegraphics[width=\linewidth]{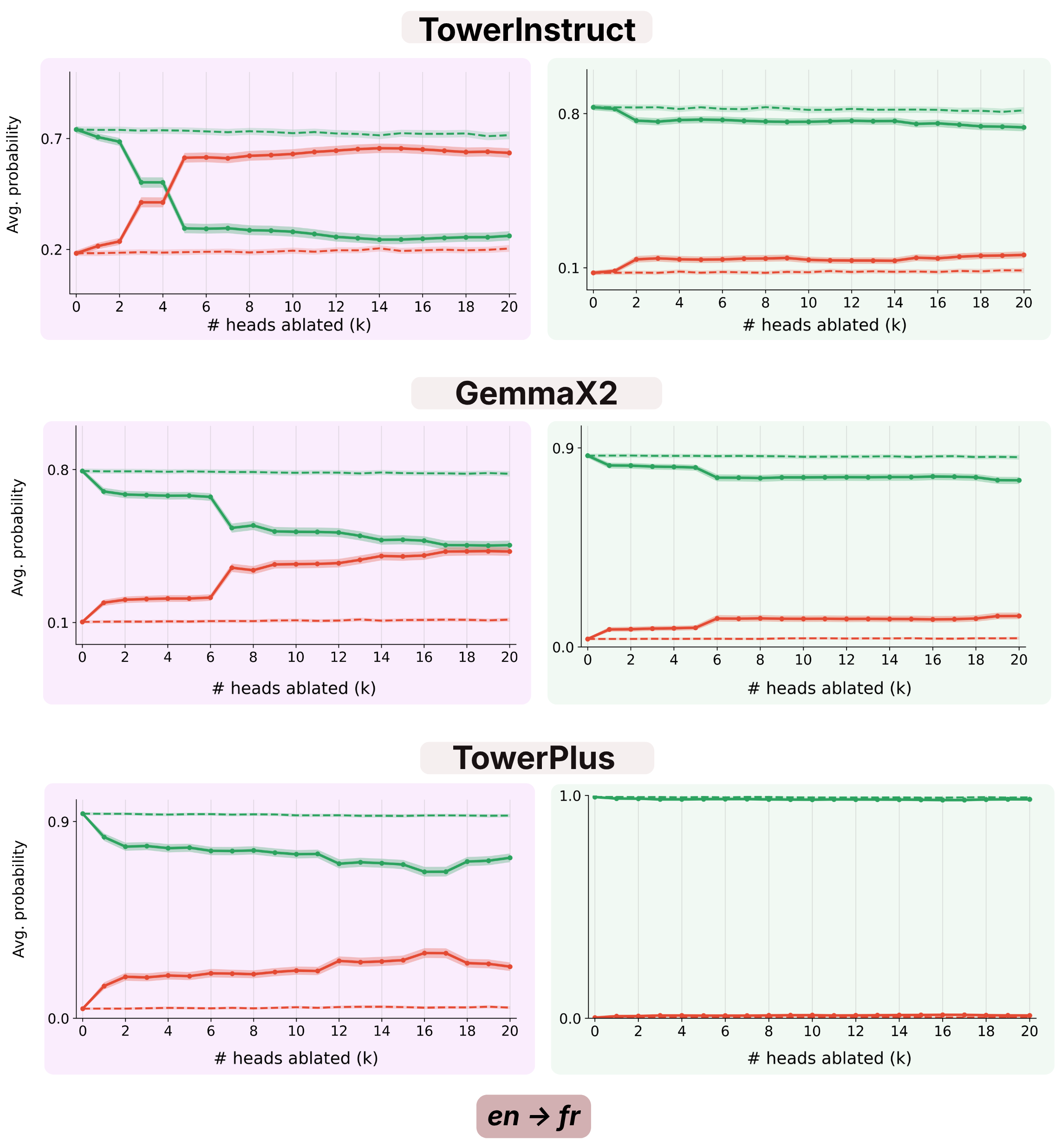}

\caption{Ablation effects for en$\rightarrow$fr across models. Panels correspond to \femalehl{feminine} and \malehl{masculine} contexts. Solid curves show the probability assigned to the correct (green) and incorrect (red) gender; dashed curves show random-ablation controls. Shaded areas denote 95\% CI across examples. Sample sizes (post filtering) are $n=398$ for TowerInstruct, $n=603$ for GemmaX2, and $n=765$ for TowerPlus.}
\label{fig:full_ablation_fr}
\end{figure*}

\begin{figure*}[h!]
\centering
\includegraphics[width=\linewidth]{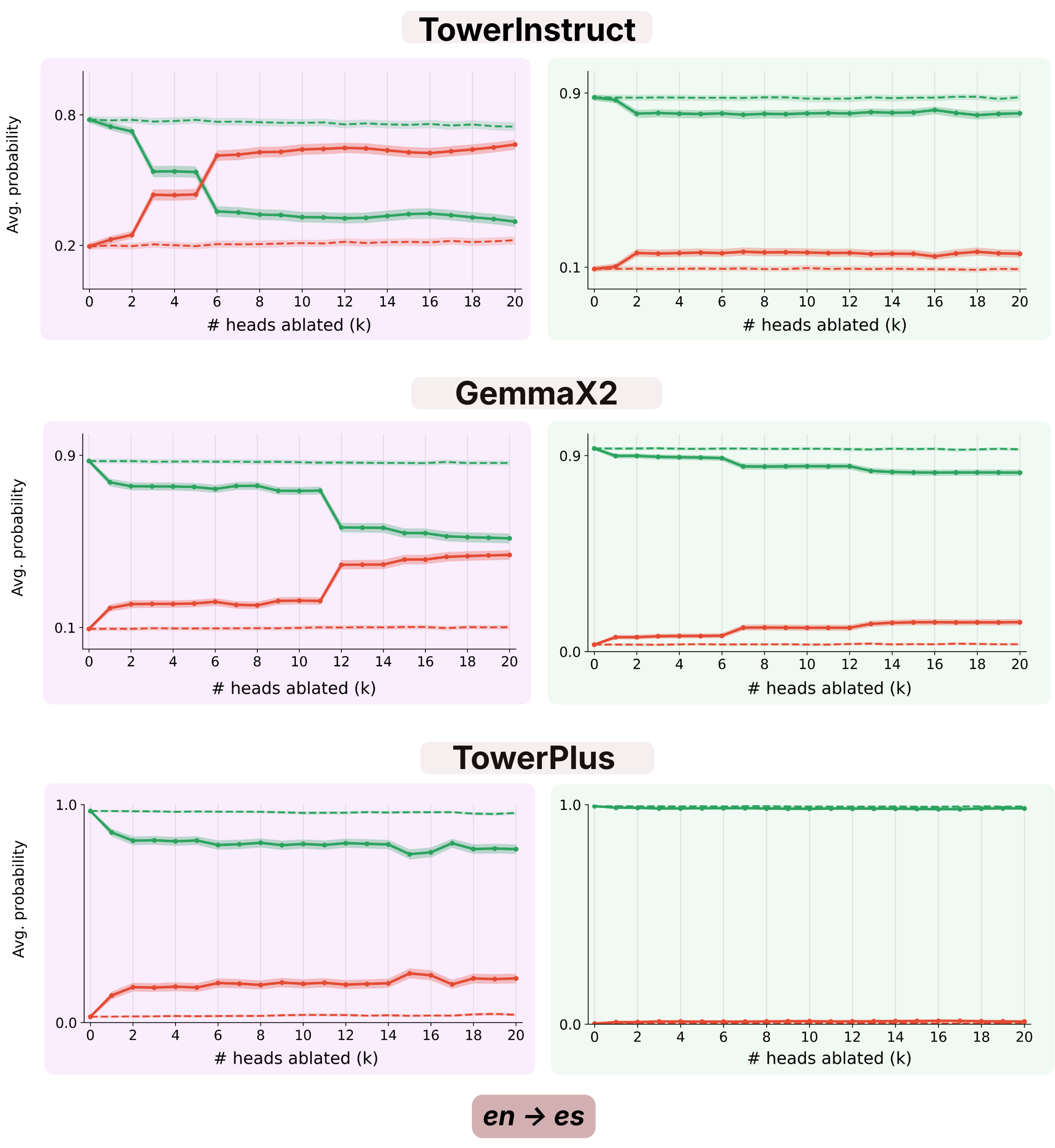}

\caption{Ablation effects for en$\rightarrow$es across models. Panels correspond to \femalehl{feminine} and \malehl{masculine} contexts. Solid curves show the probability assigned to the correct (green) and incorrect (red) gender; dashed curves show random-ablation controls. Shaded areas denote 95\% CI across examples. Sample sizes (post filtering) are $n=336$ for TowerInstruct, $n=521$ for GemmaX2, and $n=656$ for TowerPlus.}
\label{fig:full_ablation_es}
\end{figure*}

Figures~\ref{fig:full_ablation_it},
\ref{fig:full_ablation_fr}, and \ref{fig:full_ablation_es} illustrate the full ablation curves for all model-language settings and both gender conditions. For each setting, we track how the probability assigned to the correct and competing gender realizations changes as increasingly many cue-sensitive heads are ablated.

\section{Flip Rates and Stereotypical Alignment}
\label{app:flip_stereotype}

We examine whether the ablation effects interact with the stereotypical alignment of gender realizations. For each gender
$g\in\{\text{feminine},\text{masculine}\}$ and condition
$c\in\{\text{Pro},\text{Anti}\}$, we compute
\[
P(\mathrm{flip}\mid g,c)
=
\frac{N_{\mathrm{flip},g,c}}{N_{g,c}},
\]
where $N_{g,c}$ is the number of eligible examples for gender $g$ in condition $c$. Rates are computed at the same $\mathbf{k}$ values reported in
Table~\ref{tab:flip_rates}.

\begin{table}[h!]
\centering
\small
\begin{tabular}{lc >{\columncolor{fem}}c >{\columncolor{fem}}c >{\columncolor{masc}}c >{\columncolor{masc}}c}
\toprule
\textbf{Model} & \textbf{$\mathbf{k}$} & \textbf{Pro} & \textbf{Anti} & \textbf{Pro} & \textbf{Anti} \\
\midrule
\rowcolor{lightblue}
\multicolumn{6}{c}{\textbf{en$\rightarrow$it}} \\
\midrule
TowerInstruct & 4  & 69.3\% & 78.5\% & 6.7\% & 12.9\% \\
GemmaX2       & 12 & 31.3\% & 69.8\% & 0.0\% & 15.6\% \\
TowerPlus     & 7  & 28.8\% & 26.1\% & 0.0\% & 1.5\% \\

\midrule
\rowcolor{lightblue}
\multicolumn{6}{c}{\textbf{en$\rightarrow$fr}} \\
\midrule
TowerInstruct & 5  & 74.2\% & 83.3\% & 1.9\% & 7.1\% \\
GemmaX2       & 17 & 24.4\% & 71.4\% & 0.6\% & 18.8\% \\
TowerPlus     & 16 & 24.4\% & 29.8\% & 0.4\% & 3.2\% \\

\midrule
\rowcolor{lightblue}
\multicolumn{6}{c}{\textbf{en$\rightarrow$es}} \\
\midrule
TowerInstruct & 6  & 66.2\% & 82.1\% & 2.2\% & 10.8\% \\
GemmaX2       & 17 & 27.5\% & 66.6\% & 0.0\% & 12.6\% \\
TowerPlus     & 15 & 20.1\% & 18.7\% & 0.0\% & 1.6\% \\
\bottomrule
\end{tabular}
\caption{Flip probability conditioned on stereotype congruence for \femalehl{feminine} and \malehl{masculine} contexts.}
\label{tab:flip_stereotype}
\end{table}

Across models and languages, \emph{Anti}-stereotypical realizations generally exhibit higher flip rates than \emph{Pro}-stereotypical ones. In masculine contexts this holds for every model and language, although rates remain low throughout. In feminine contexts the difference is largest for GemmaX2, more moderate for TowerInstruct, negligible for TowerPlus.

\FloatBarrier
\clearpage

\FloatBarrier
\clearpage

\FloatBarrier
\clearpage

\providecommand{\attnpanel}{}
\renewcommand{\attnpanel}[2]{%
  \includegraphics[width=\textwidth,height=#2\textheight,keepaspectratio]{#1}\par\medskip}

\clearpage
\begin{figure*}[p]
\centering
\attnpanel{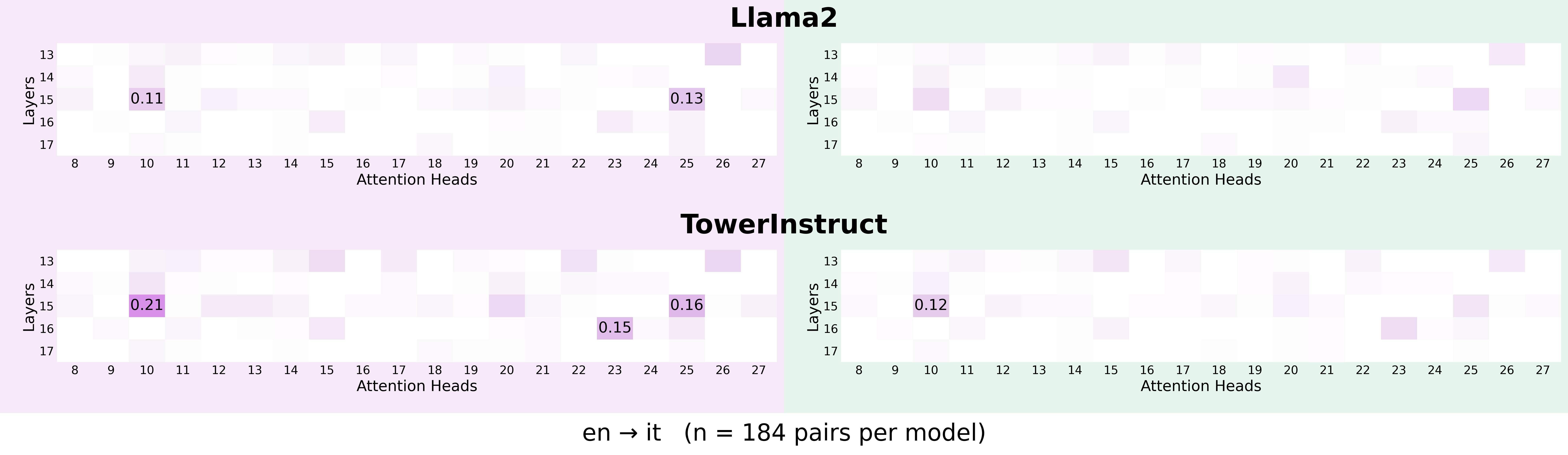}{0.2}
\attnpanel{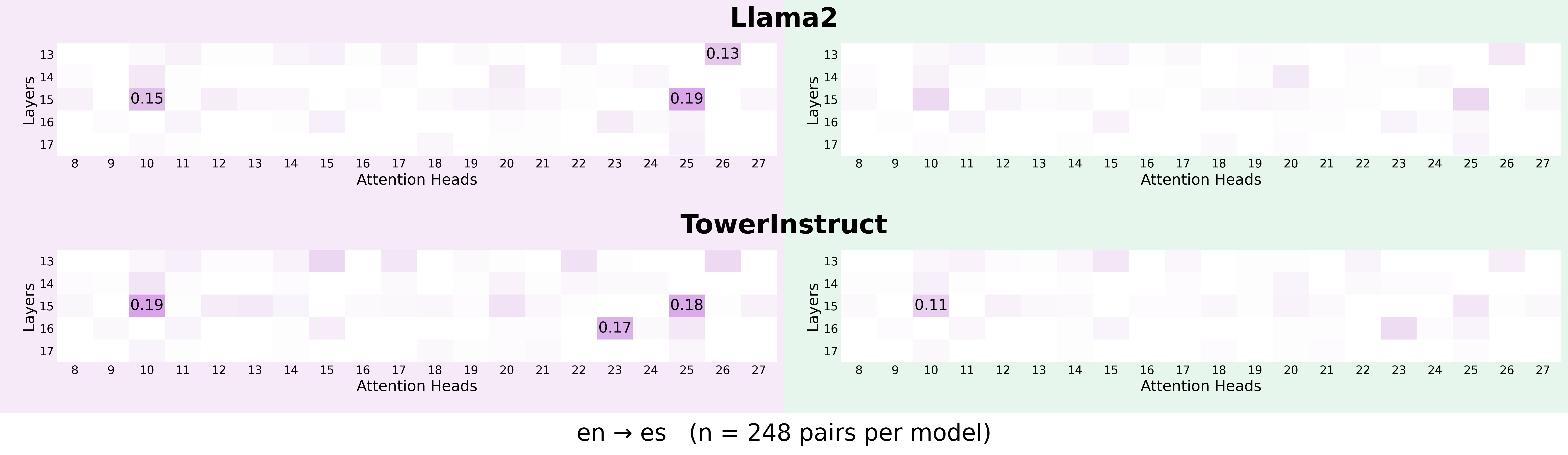}{0.2}
\attnpanel{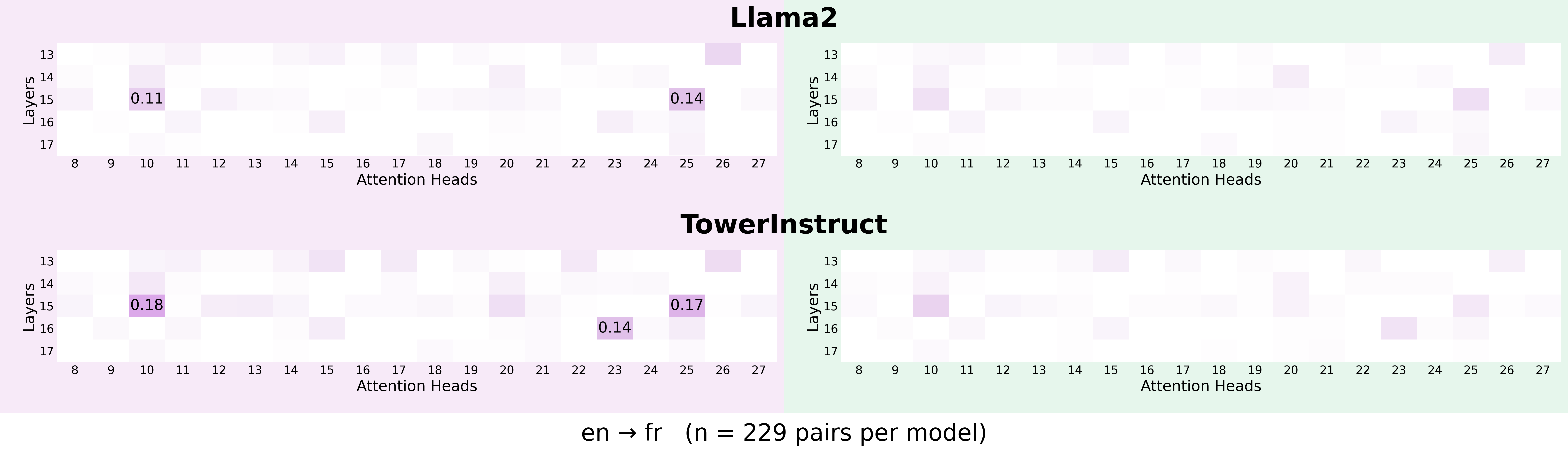}{0.2}
\attnpanel{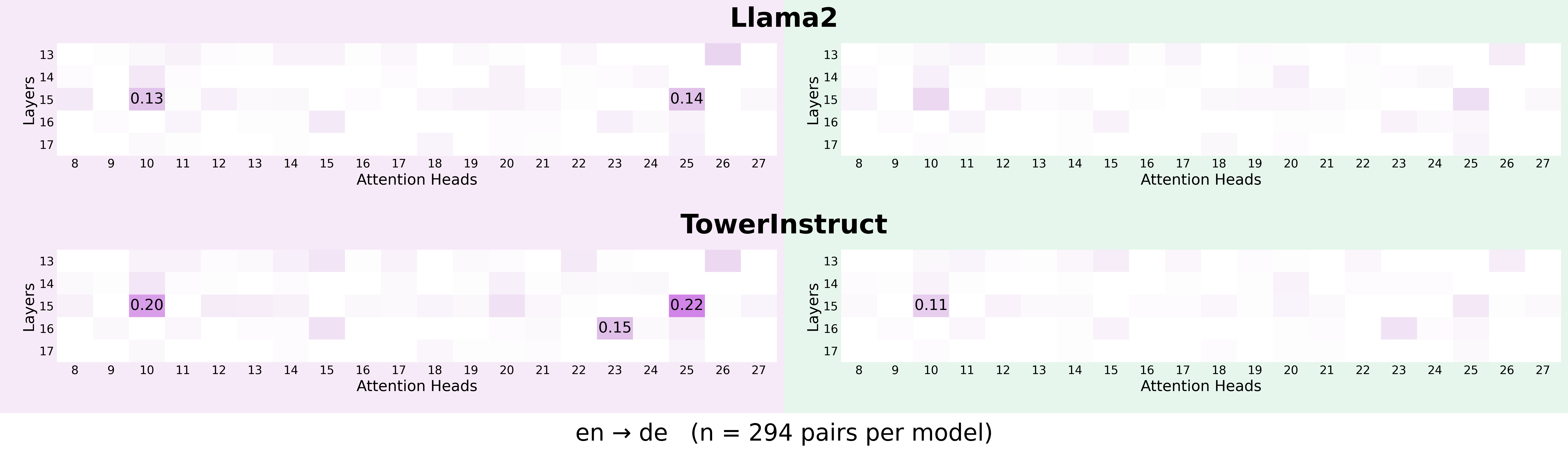}{0.2}
\caption{Snapshot of attention to the \femalehl{feminine} vs.\ \malehl{masculine} cue in Llama2-based models over
equally sized subsets of correctly resolved minimal pairs ($n_{it}=184$, $n_{es}=248$, $n_{fr}=229$, $n_{de}=294$).
Layers are shown on the $y$-axis and heads on the $x$-axis. Most prominent scores are annotated.}
\label{fig:appendix_llama2_correct}
\end{figure*}

\clearpage
\begin{figure*}[p]
\centering
\attnpanel{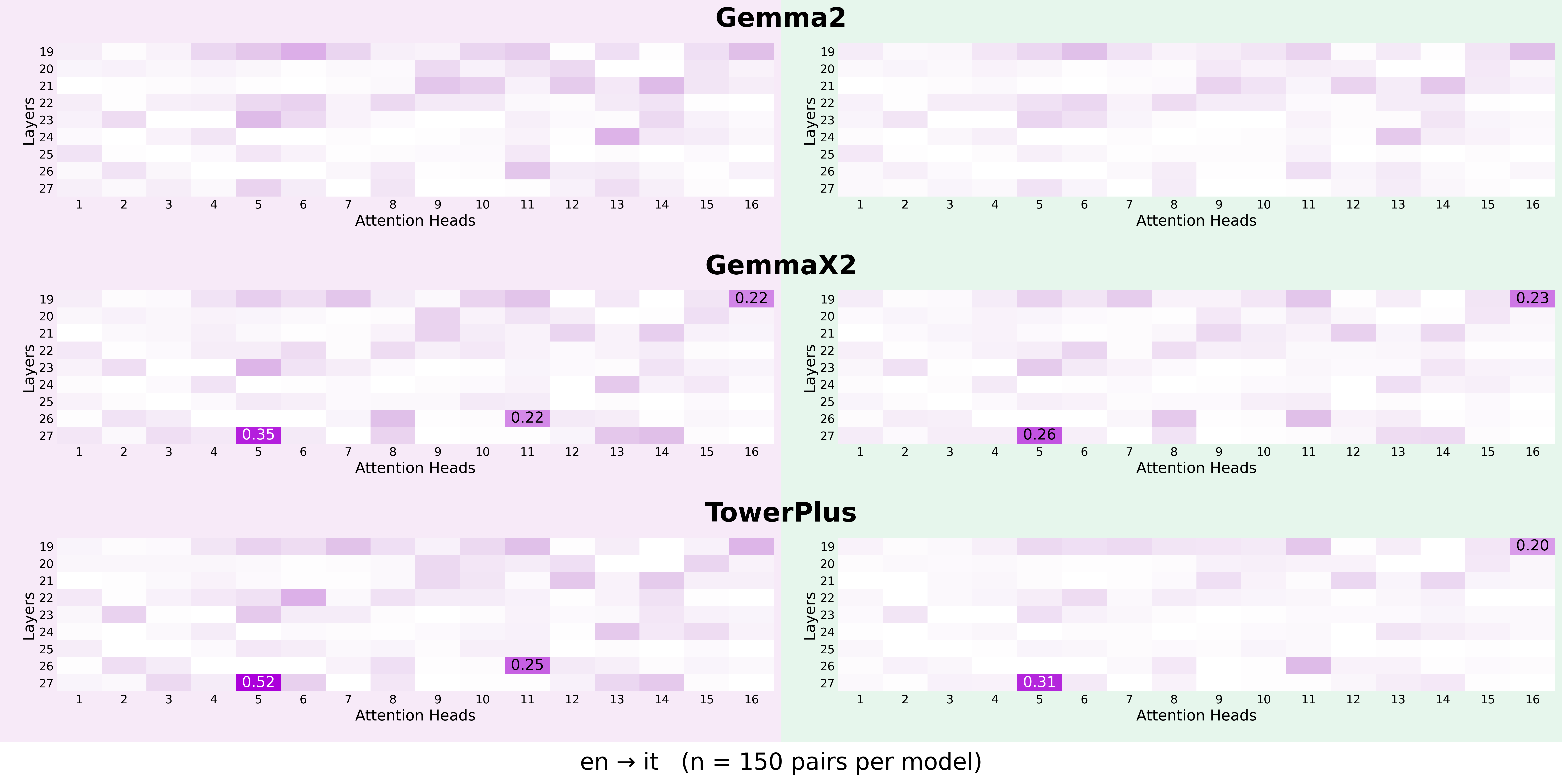}{0.42}
\attnpanel{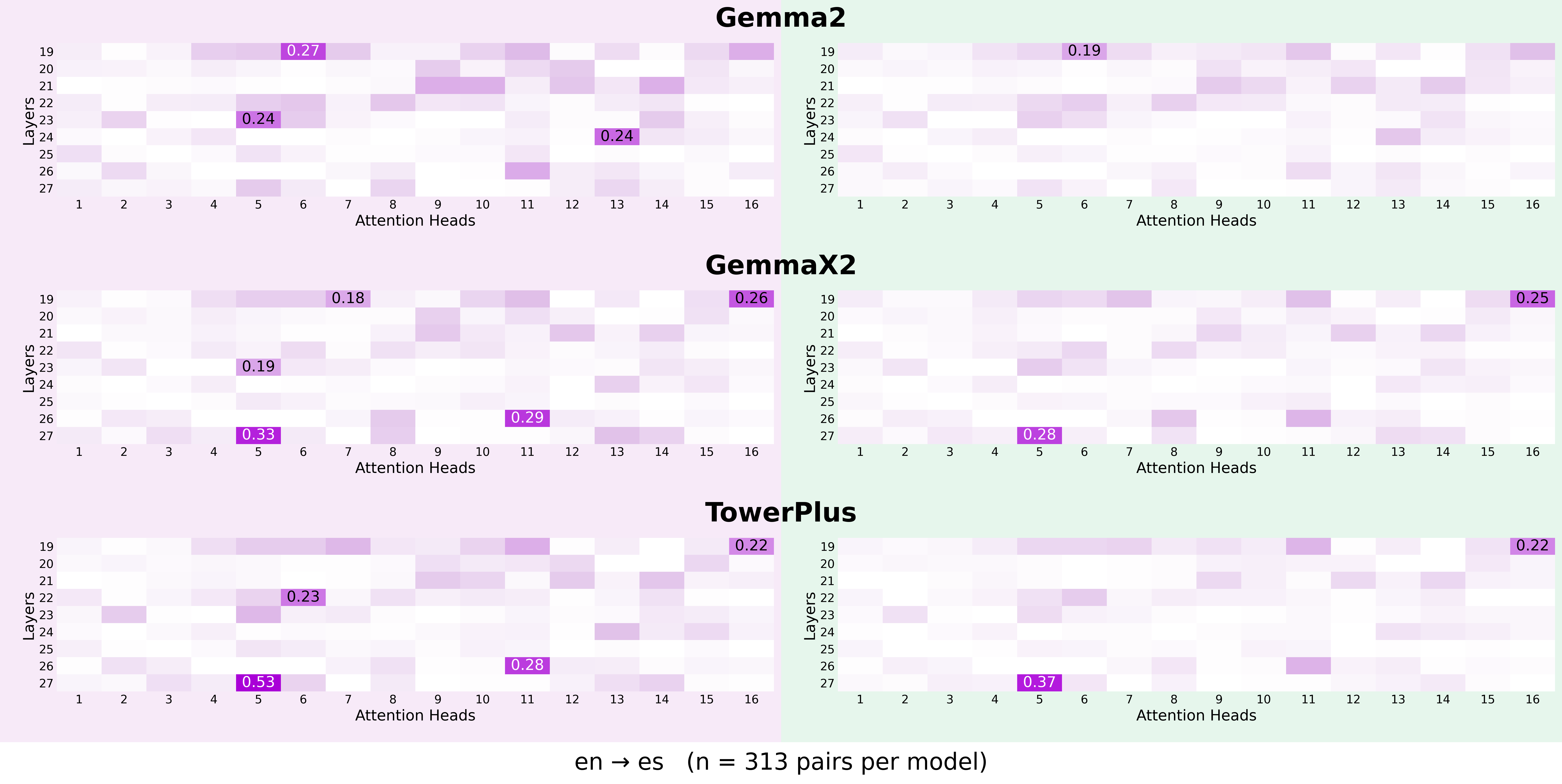}{0.42}
\caption{Snapshot of attention to the \femalehl{feminine} and \malehl{masculine} cues in Gemma2-based models over
equally sized subsets of correctly resolved minimal pairs ($n_{it}=150$, $n_{es}=313$, $n_{fr}=155$, $n_{de}=364$).
Layers are shown on the $y$-axis and heads on the $x$-axis. Scores $\geq 0.18$ are annotated.
Continued on the next page.}
\label{fig:appendix_gemma2_correct}
\end{figure*}

\begin{figure*}[p]
\ContinuedFloat
\centering
\attnpanel{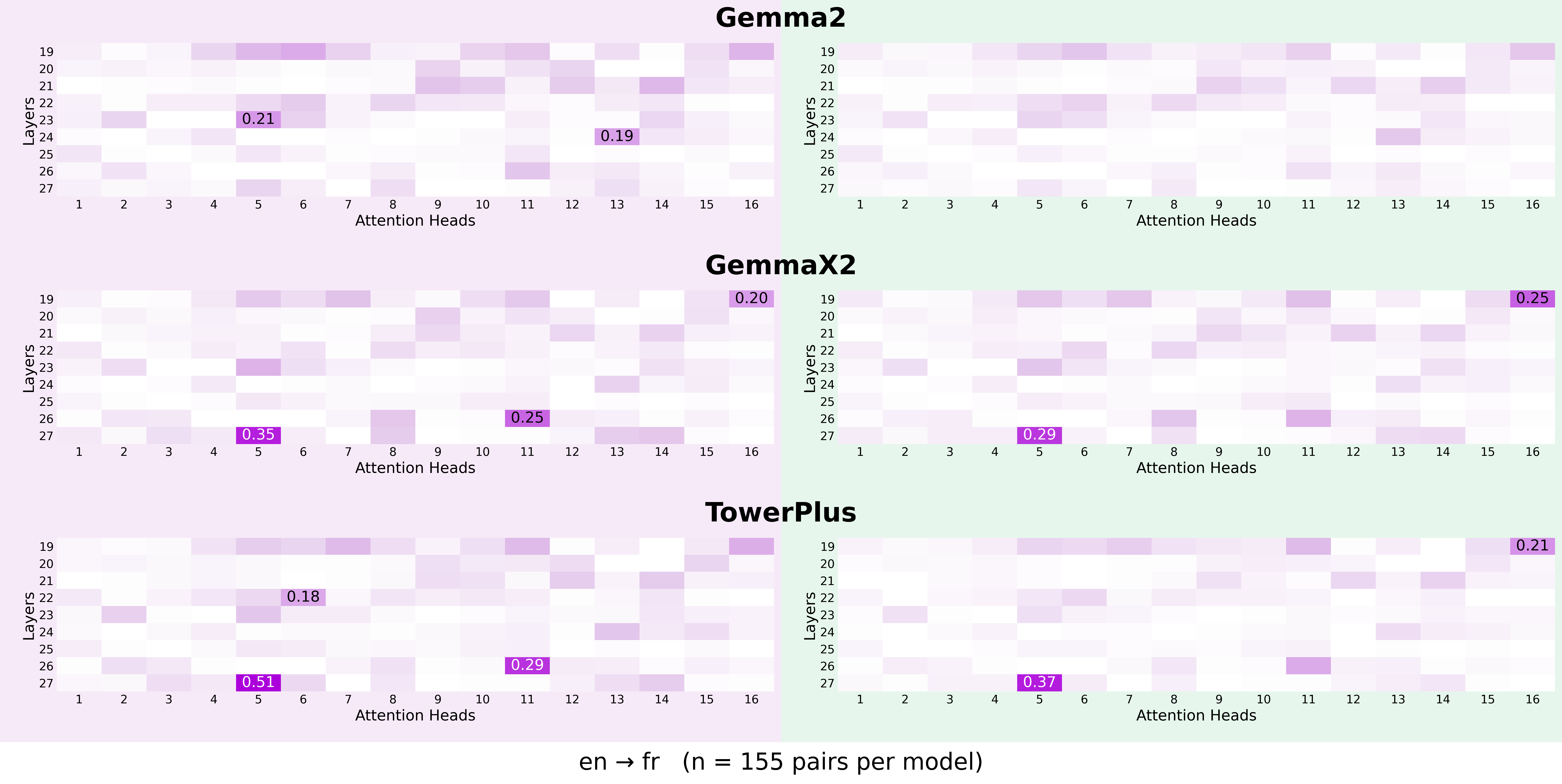}{0.42}
\attnpanel{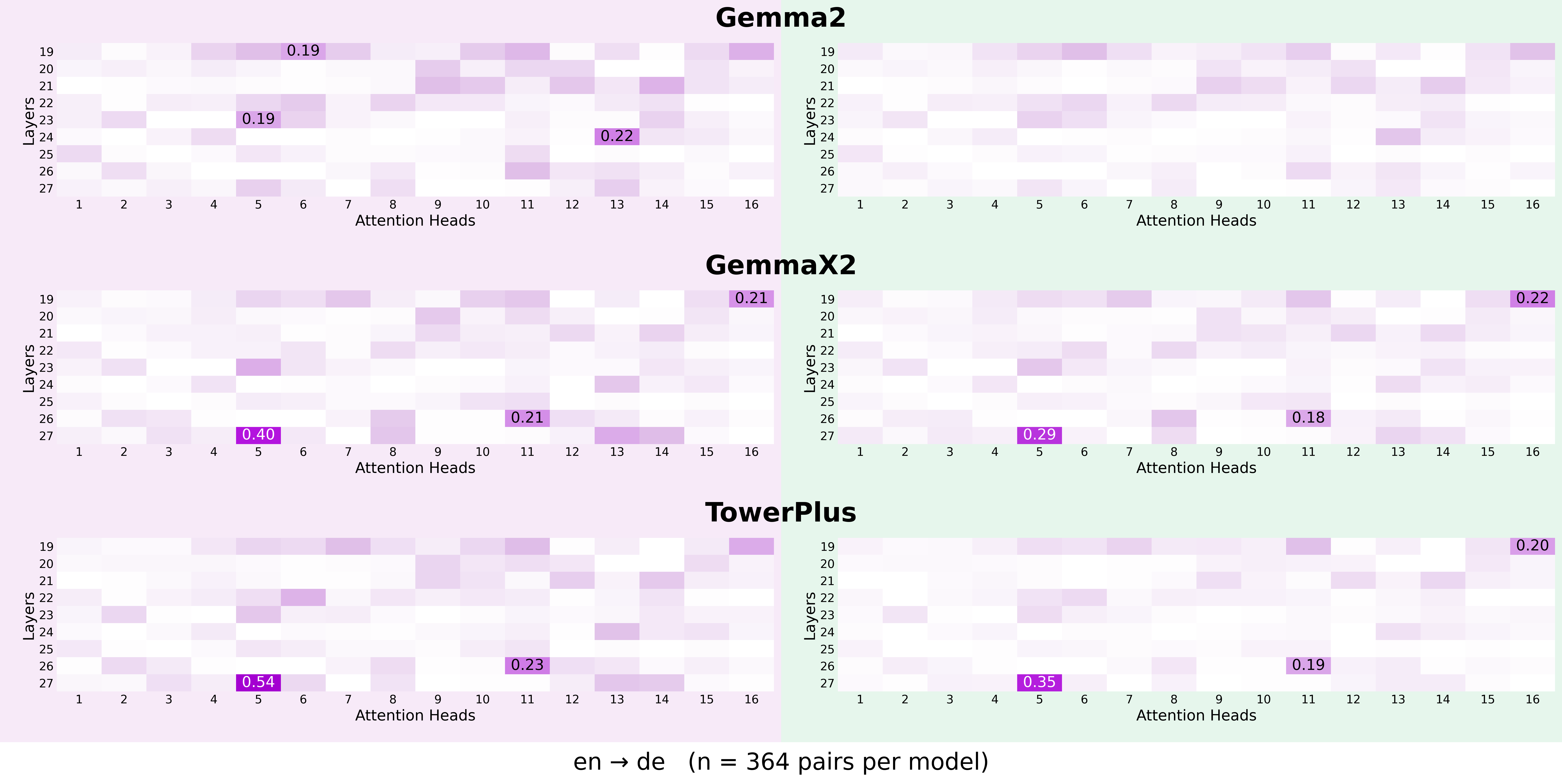}{0.42}
\caption[]{(continued) Gemma2-based models, en$\rightarrow$fr and en$\rightarrow$de.}
\end{figure*}

\FloatBarrier
\clearpage

\section{On the Validation of Article-Based Gender Probabilities}\label{app:articles}
To validate the choice of focusing on article-based probabilities in Section~5.2, we compute the agreement between the grammatical gender expressed by the (preceding) article and that of the corresponding profession noun in Italian, Spanish and French translation outputs. For each (target) language, we construct a dictionary of masculine and feminine profession realizations by combining lexical resources, such as BabelNet \citep{navigli-ponzetto-2010-babelnet}, with manual inspection of model outputs. Table~\ref{tab:profession_dictionary_examples2} provides some illustrative examples.

\begin{table}[h!]
\centering
\small
\setlength{\tabcolsep}{4pt}
\renewcommand{\arraystretch}{1.05}
\resizebox{\columnwidth}{!}{
\begin{tabular}{ll}
\toprule
\textbf{Profession} & \textbf{Variants} \\
\midrule

\multirow{9}{*}{\textbf{
teacher}}
& \textbf{Italian} \\
 & maestro, professore, istruttore \malehl{masc}\\
& maestra, professoressa, istruttrice \femalehl{fem}\\
\cmidrule(lr){2-2}
& \textbf{Spanish} \\
& maestro, profesor, instructor \malehl{masc}\\
& maestra, profesora, instructora \femalehl{fem}\\
\cmidrule(lr){2-2}
& \textbf{French} \\
& professeur, enseignant, instituteur \malehl{masc}\\
& professeure, enseignante, institutrice \femalehl{fem}\\

\midrule

\multirow{9}{*}{\textbf{manager}}
& \textbf{Italian} \\
& direttore, dirigente, amministratore \malehl{masc}\\
& direttrice, dirigente, amministratrice \femalehl{fem}\\
\cmidrule(lr){2-2}
& \textbf{Spanish} \\
& director, gerente, administrador \malehl{masc}\\
& directora, gerente, administradora \femalehl{fem}\\
\cmidrule(lr){2-2}
& \textbf{French} \\
& directeur, gérant, administrateur \malehl{masc}\\
& directrice, gérante, administratrice \femalehl{fem}\\

\midrule

\multirow{9}{*}{\textbf{writer}}
&\textbf{Italian} \\
& autore, scrittore, giornalista \malehl{masc}\\
& autrice, scrittrice, giornalista \femalehl{fem}\\
\cmidrule(lr){2-2}
&\textbf{Spanish} \\
& autor, escritor, periodista \malehl{masc}\\
& autora, escritora, periodista \femalehl{fem}\\
\cmidrule(lr){2-2}
& \textbf{French} \\
&  auteur, écrivain, journaliste \malehl{masc}\\
& autrice, écrivaine, journaliste \femalehl{fem}\\

\bottomrule
\end{tabular}
}
\caption{Examples of lexical variants from the Italian, Spanish, and French profession dictionaries used to validate article-noun gender agreement. Epicene profession nouns are retained when the preceding article provides an unambiguous gender realization (e.g., \textit{le/la journaliste} or \textit{el/la periodista}).}
\label{tab:profession_dictionary_examples2}
\end{table}

\newpage
We identify the (preceding) gender-marked article and compare its grammatical gender with that of the detected profession form. We explicitly exclude gender-invariant forms as well as cases in which either the profession noun or a valid preceding article cannot be reliably identified (see Coverage).

We observe a near-perfect agreement across all models and languages, ranging from 97.4\% to 100.0\% (Table~\ref{tab:article_agreement}), which supports the use of article-level probabilities as a reliable proxy for gender preference in our causal analysis.

\begin{table}[h!]
\centering
\small
\setlength{\tabcolsep}{4pt}
\renewcommand{\arraystretch}{1.05}
\resizebox{\columnwidth}{!}{
\begin{tabular}{lc >{\columncolor{masc}}c >{\columncolor{fem}}cc}
\toprule
\textbf{Model} & \textbf{Overall} & \textbf{Masc.} &
\textbf{Fem.} & \textbf{Coverage} \\
\midrule
\rowcolor{lightblue}
\multicolumn{5}{c}{\textbf{en$\rightarrow$it}} \\
\midrule
TowerInstruct & 97.4\% & 99.9\% & 94.8\% & 66.9\% \\
GemmaX2       & 99.3\% & 100.0\% & 98.5\% & 64.7\% \\
TowerPlus     & 98.2\% & 100.0\% & 96.3\% & 68.5\% \\
\midrule
\rowcolor{lightblue}
\multicolumn{5}{c}{\textbf{en$\rightarrow$es}} \\
\midrule
TowerInstruct & 99.5\% & 100.0\% & 99.2\% & 64.3\% \\
GemmaX2       & 100.0\% & 99.9\% & 100.0\% & 70.0\% \\
TowerPlus     & 99.4\% & 98.6\% & 99.7\% & 68.1\% \\
\midrule
\rowcolor{lightblue}
\multicolumn{5}{c}{\textbf{en$\rightarrow$fr}} \\
\midrule
TowerInstruct & 99.2\% & 100.0\% & 98.5\% & 56.5\% \\
GemmaX2       & 98.9\% & 100.0\% & 98.0\% & 61.1\% \\
TowerPlus     & 100.0\% & 100.0\% & 100.0\% & 60.9\% \\

\bottomrule
\end{tabular}
}
\caption{Agreement between the grammatical gender expressed by the (preceding) article and the corresponding profession noun in Italian, Spanish, and French translations. Results are reported overall and separately for \malehl{masculine} and \femalehl{feminine} realizations.}
\label{tab:article_agreement}
\end{table}

\end{document}